\documentclass[preprint,3p, 12pt]{elsarticle}
\usepackage{hyperref}
\hypersetup{hidelinks,
	colorlinks=true,
	allcolors=black,
	pdfstartview=Fit,
	breaklinks=true}

\usepackage{amsmath}
\usepackage{amssymb}
\usepackage{multirow}
\usepackage{adjustbox}
\usepackage{makecell}
\usepackage{subfig}
\usepackage{overpic}  
\usepackage{float}
\biboptions{numbers,sort&compress}

\usepackage{lineno}

\journal{ISPRS Journal of Photogrammetry and Remote Sensing}

\begin{document}

\begin{frontmatter}



\title{LMFNet: An Efficient Multimodal Fusion Approach for Semantic Segmentation in High-Resolution Remote Sensing}


\author[inst1]{Tong Wang}
\author[inst1]{Guanzhou Chen\corref{cor1}}
\author[inst1]{Xiaodong Zhang\corref{cor1}}
\affiliation[inst1]{organization={State Key Laboratory of information Engineering in Surveying, Mapping and Remote Sensing, Wuhan University},
            addressline={299 Bayi Road, Wuchang District}, 
            city={Wuhan},
            postcode={430072}, 
            state={Hubei Province},
            country={China}}

\author[inst2]{Chenxi Liu}
\author[inst1]{Jiaqi Wang}
\author[inst1]{Xiaoliang Tan}
\author[inst1]{Wenlin Zhou}
\author[inst1]{Chanjuan He}

\affiliation[inst2]{organization={Electronic Information School, Wuhan University},
            addressline={299 Bayi Road, Wuchang District}, 
            city={Wuhan},
            postcode={430072}, 
            state={Hubei Province},
            country={China}}

\cortext[cor1]{ Corresponding authors: cgz@whu.edu.cn (G. Chen), zxdlmars@whu.edu.cn (X. Zhang).}

\begin{abstract}
Despite the rapid evolution of semantic segmentation for land cover classification in high-resolution remote sensing imagery, integrating multiple data modalities such as Digital Surface Model (DSM), RGB, and Near-infrared (NIR) remains a challenge. Current methods often process only two types of data, missing out on the rich information that additional modalities can provide. Addressing this gap, we propose a novel \textbf{L}ightweight \textbf{M}ultimodal data \textbf{F}usion \textbf{Net}work (LMFNet) to accomplish the tasks of fusion and semantic segmentation of multimodal remote sensing images. LMFNet uniquely accommodates various data types simultaneously, including RGB, NirRG, and DSM, through a weight-sharing, multi-branch vision transformer that minimizes parameter count while ensuring robust feature extraction.  Our proposed multimodal fusion module integrates a \textit{Multimodal Feature Fusion Reconstruction Layer} and \textit{Multimodal Feature Self-Attention Fusion Layer}, which can reconstruct and fuse multimodal features. Extensive testing on public datasets such as US3D, ISPRS Potsdam, and ISPRS Vaihingen demonstrates the effectiveness of LMFNet. Specifically, it achieves a mean Intersection over Union ($mIoU$) of 85.09\% on the US3D dataset, marking a significant improvement over existing methods. Compared to unimodal approaches, LMFNet shows a 10\% enhancement in $mIoU$ with only a 0.5M increase in parameter count. Furthermore, against bimodal methods, our approach with trilateral inputs enhances $mIoU$ by 0.46 percentage points. Our LMFNet is not only scalable and accurate but also efficient in terms of parameter usage, indicating its potential for broad application in remote sensing imagery analysis. The proposed network paves the way for more effective and efficient integration of multiple data modalities, promising significant advancements in land cover classification tasks.
\end{abstract}


\begin{highlights}
    \item We propose a novel lightweight multimodal remote sensing data fusion and semantic segmentation network: LMFNet.
   \item The LMFNet we proposed can accept inputs from any number of modalities, exhibiting high scalability and flexibility.
    \item We propose a novel fusion module that includes a Multimodal Feature Fusion Reconstruction Layer and a Multimodal Feature Self-Attention Fusion Layer.
    \item An in-depth analysis of the changes in multi-scale and multimodal features during the data fusion process is carried out.
\end{highlights}

\begin{keyword}
Multimodal Data Fusion \sep Remote Sensing \sep Lightweight Network \sep Semantic Segmentation \sep Digital Surface Model
\PACS 0000 \sep 1111
\MSC 0000 \sep 1111
\end{keyword}

\end{frontmatter}


\section{Introduction}
\label{sec:intro}

Semantic segmentation of land cover on Very High Resolution (VHR) remote sensing data is a primary application domain of remote sensing technology~\cite{chen2018symmetrical,li_deep_2022,paoletti_deep_2019,wang_smilies_2023,ma_deep_2019}. This task is enhanced by integrating diverse types of data, including multispectral imagery such as RGB and Near-infrared (NIR), and Digital Surface Model (DSM) data. Together, these data sources provide a comprehensive foundation for land cover classification from the aspects of spectrum, texture, and three-dimensional structure~\cite{zhang_deep_2021,zhang_advances_2017,liao_s_2023,liao_linear_2022,zhang_distinguishable_2022,liu_dense_2020,li_deep_2022,parajuli_fusion_2018,cai_detecting_2023,zhang_deep_2023}. 
Although deep learning-based semantic segmentation technology has made great progress, leveraging methods such as FCN~\cite{long_fully_2015}, SegNet~\cite{segnet}, SDFCN~\cite{chen_sdfcnv2_2021}, UNet~\cite{navab_u-net_2015}, PSPNet~\cite{zhao2017pyramid}, DeeplabV3+~\cite{chen2018encoder}, HRNet~\cite{wang2020deep}, and Segformer~\cite{xie2021segformer}, there are still practical challenges in effectively fusing RGB, NIR, and DSM data to enhance land cover classification~\cite{aleissaee_transformers_2023,huang_automatic_2019,lai_semantic_2017,zhang_deep_2021}.



In early studies, researchers used machine learning methods and manually extracted features for fusion classification~\cite{singh_lidar-landsat_2012, pal_random_2005,zhang2018exploring}. These methods, including the application of Support Vector Machines (SVM)~\cite{fierrez-aguilar_fusion_2003,zhang2018exploring}, and Random Forests (RF)~\cite{ke_synergistic_2010,gonzalez_-board_2016}, showcased potential by combining multimodal features~\cite{bagui_combining_2005}. However, the increased dimensionality of multimodal features often led to the curse of dimensionality~\cite{li_deep_2022}. To this end, Principal Component Analysis (PCA), Independent Component Analysis (ICA), and Sparse Low-Rank Component Analysis (SLCA)~\cite{rasti2017fusion} have been proposed to reduce the dimension of fusion features. Furthermore, various shape, geometric features such as Extended Attribute Profile Contours (EAP)~\cite{pedergnana2012classification}, Attribute Profiles~\cite{ghamisi2014fusion}, etc., have also been introduced for multimodal fusion and classification. Yet, the aforementioned manual approaches struggled to extract deep and non-linear features, yielded limited performance enhancements.

\begin{figure}[htbp]

\centering
	\subfloat[Data-level]{\label{fig:datalevel}
 \includegraphics[width=.17\columnwidth, height=59mm]{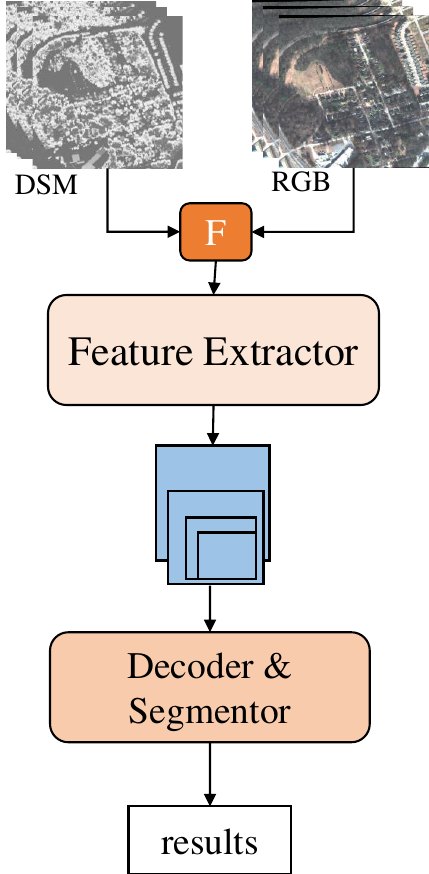}}\hspace{5pt}
	\subfloat[Decision-level]{\label{fig:dclevel}
 \includegraphics[width=.21\columnwidth, height=59mm]{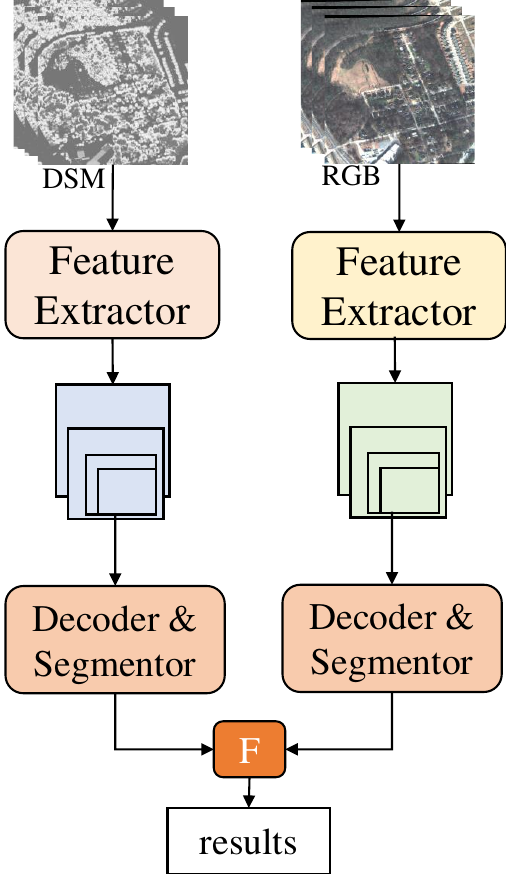}}\hspace{5pt}
	\subfloat[Feature-level]{\label{fig:mfusion}
 \includegraphics[width=.22\columnwidth, height=59mm]{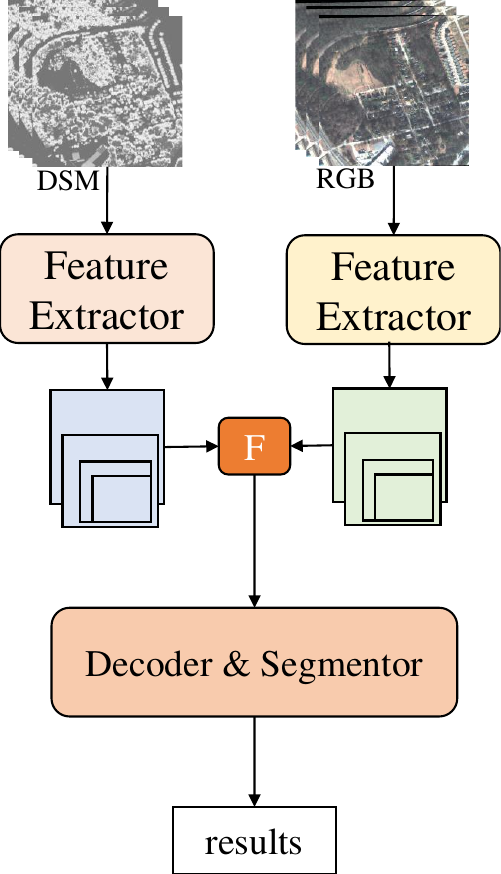}}\hspace{5pt}
	\subfloat[Ours]{\label{fig:ours}
 \includegraphics[width=.27\columnwidth, height=59mm]{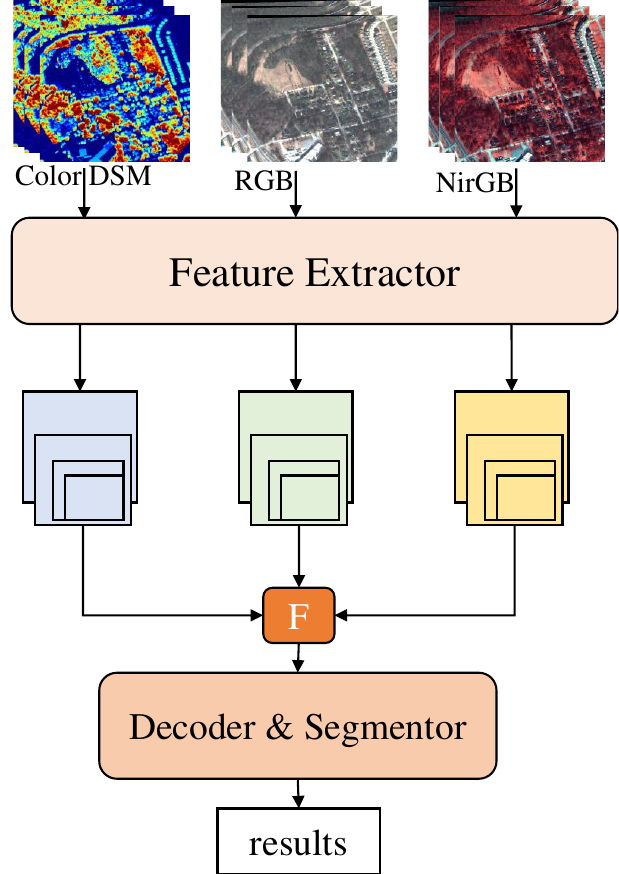}}
	\caption{\label{fig:inro}Comparison of different fusion frameworks. In the figure, \textbf{F} represents the fusion module or fusion method.}
\end{figure}

The development of deep learning techniques, especially Convolutional Neural Networks (CNN) and Vision-Transformers (ViT), has greatly improved the feature extraction and fusion process, by  adaptively and automatically modeling the complex relationships between input and output data~\cite{abdollahi_deep_2020,ghamisi_multisource_2019,han_multimodal_2022,song_joint_2024,zhang_object-based_2022}. The current deep learning based data fusion and semantic segmentation methods can be divided into \textbf{data-level}, \textbf{decision-level}, and \textbf{feature-level fusion}~\cite{zhang_deep_2021,li_deep_2022,hosseinpour_cmgfnet_2022}, as shown in Figure~\ref{fig:inro}. \textbf{Data-level} fusion (as shown in Figure~\ref{fig:datalevel}) merges multimodal data through methods such as channel concatenation and pansharpening, and then inputs it into the network~\cite{hosseinpour_cmgfnet_2022}. However, the simple concatenation of images provides limited help for the extraction of multimodal features, and the high variability of DSM (or depth map in Computer Vsion) to some extent increases the uncertainty of feature learning~\cite{li_deep_2022}. \textbf{Decision-level} fusion (as shown in Figure~\ref{fig:dclevel}) adopts two different methods for classification and fuses the prediction results at the end~\cite{mohammadi2020object}. This fusion method can provide more scalability and flexibility than earlier fusion methods, but the complementary features between multiple modalities are discarded~\cite{li_deep_2022}. 

In the field of \textbf{feature-level} fusion (as shown in Figure~\ref{fig:mfusion}) research, the technological development history reflects an evolution from basic concatenation fusion methods (FuseNet~\cite{lai_fusenet_2017}, vFuseNet~\cite{audebert2018beyond}, \cite{wu2020depth}) to more complex fusion techniques. These include fusions that introduce spatial and channel attention based fusion techniques (HAFNet~\cite{zhang2020hybrid}, NLFNet~\cite{yan2021nlfnet}) as well as methods that utilize gated weighting fusion techniques (SA-Gate~\cite{chen2020bi}, CMGAN\cite{liu2020cross}, ABMDRNet~\cite{zhang2021abmdrnet}). In recent years, attention matrix exchange fusion methods~\cite{zhang2023cmx} based on self-attention mechanisms and ViT~\cite{dosovitskiy_image_2021} technology have further expanded this field. These developments have significantly enhanced the expressive capabilities of fused features, especially in making breakthrough progress in dealing with complex interactions related to spatial alignment and channel correlation across various data modalities~\cite{li_deep_2022,hosseinpour_cmgfnet_2022}. 
Despite significant technological advancements, current research has not fully explored the complexities associated with multimodal feature fusion mechanisms, primarily focusing on the fusion of bimodal data. Specifically, the fusion of NIR, RGB, and DSM data in semantic segmentation tasks presents a notable challenge due to the lack of frameworks adept at merging these diverse data types effectively.


To address the issue of insufficient utilization of RGB, Nir and DSM data , we propose the  \textbf{L}ightweight \textbf{M}ultimodal data \textbf{F}usion \textbf{Net}work, namely, LMFNet, as shown in Figure~\ref{fig:ours}, designed to streamline the fusion process through innovative feature reconstruction and fusion strategies. The main contributions of this paper include: 
 \begin{itemize}
    \item We design a novel multimodal feature fusion module (MFM) to achieve seamlessly fusion of arbitrarily multiple modal data without the need to modify the fusion module's structure.
   \item We propose a comprehensive multimodal fusion and semantic segmentation network, LMFNet, which incorporates MFM as well as a preprocessing pipeline for multimodal data.
    \item Through experiments conducted on three datasets, we conduct an in-depth analysis of the internal fusion principles of the proposed fusion module and the details of fusion before and after the multimodal feature integration.
\end{itemize}

 The structure of the paper is organized as follows: In section~\ref{sec:method}, we provide a detailed introduction to the structure of the proposed LMFNet and the multimodal data fusion method, with some data pre-processings also introduced at the end of this section. In section~\ref{sec:exps}, the datasets, experimental methods, and results are introduced, followed by an analysis and discussion. The advantages and disadvantages of our method are discussed in section~\ref{sec:discusiion}. Finally, a summary is provided in section~\ref{sec:con}, along with plans for future research.

\section{Method}
\label{sec:method}

\begin{figure}[ht]
    \centering
    \includegraphics[width=1\linewidth]{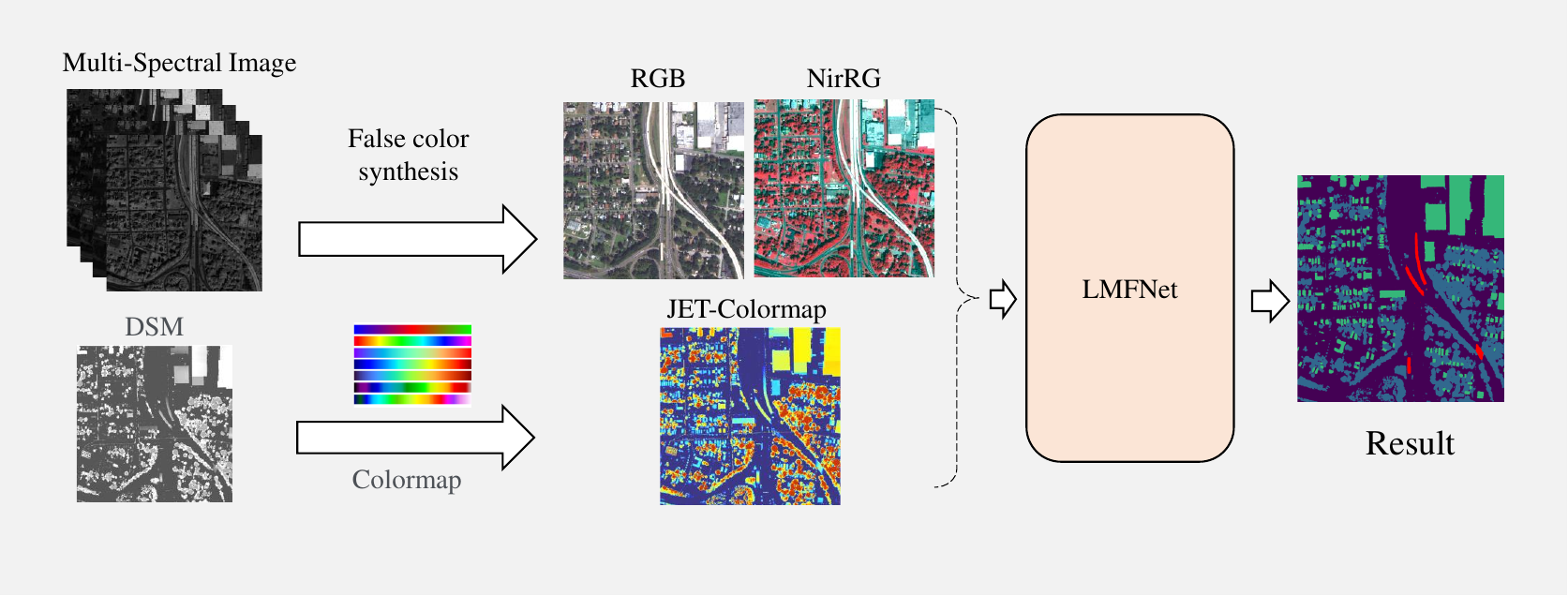}
    \caption{Overall framework. To facilitate the input of data into the network, multi-spectral data is separated into RGB and NirRG through false color synthesis; a single-band DSM is converted into RGB by using colormap. }
    \label{fig:dataframe}
\end{figure}

The advantage of the heterogeneity among RGB, NirRG data and DSM data lies in the complementary information \cite{huang_automatic_2019,ramachandram_deep_2017}, However, each time a new modality is added, the network and the entire framework need to be redesigned. To address this issue, we design the framework as shown in Figure~\ref{fig:dataframe}. For multi-spectral data, we adapt the method of false-color synthesis to obtain three-band data. For single-band data like DSM, we employ a colormap-based rendering approach to generate three-band data. Our specially designed LMFNet is capable of processing data from any number of modalities without requiring modifications to its architecture.


\begin{figure}[ht]
    \centering
    \includegraphics[width=\linewidth]{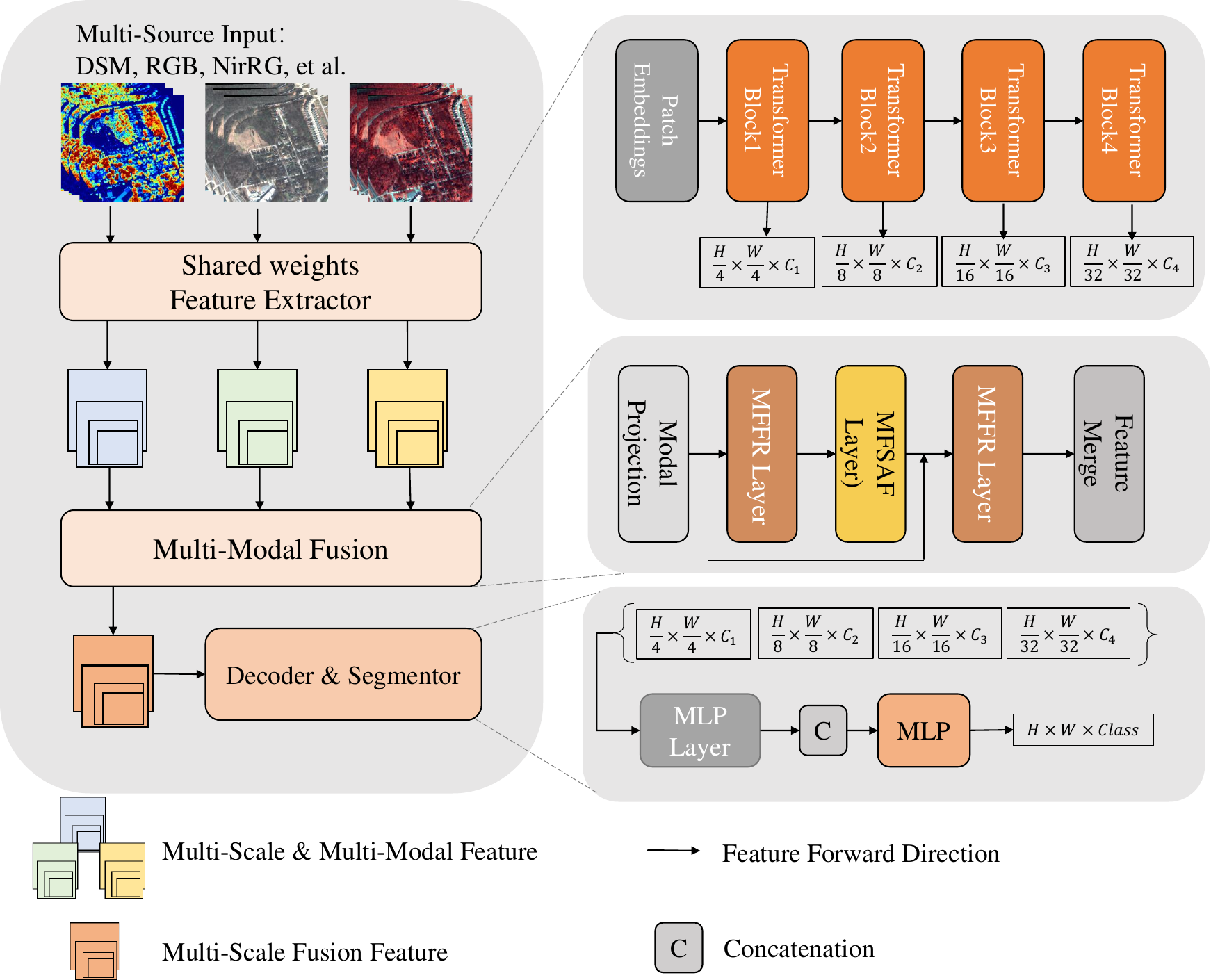}
    \caption{ LMFNet structure. On the left side of the figure is the overall structure, on the upper right side is the structure of the encoder, in the middle right is the structure of the multimodal fusion module, and the bottom right shows the structure of the decoder and classifier. }
    \label{fig:framework}
\end{figure}

The Figure~\ref{fig:framework} illustrates the proposed LMFNet. For the input multimodal data, a weight sharing multi-branch backbone network is used to extract multi-scale features. Subsequently, the features are fused using our proposed multimodal feature fusion module and then fed into an Multilayer-Perceptron (MLP) decoder for prediction and obtaining the final classification results. The following will first introduce the overall structure of the framework and then detail each module separately.

\subsection{LMFNet architecture}
\label{sec:netarch}
Previous studies on feature extraction employ a diverse array of heterogeneous architectures to facilitate the extraction of multimodal features~\cite{zhang_deep_2021,li_deep_2022,bittner_building_2018}. This has led to a significant increase in the number of parameters. This work introduces a novel network with shared-weights multimodal feature extractor, thereby significantly reducing the total number of parameters in the network. The feature extractor takes inspiration from the architectural philosophy of MiT~\cite{xie2021segformer}, ensuring a balance between efficiency and advancement, as shown in the upper right part of Figure~\ref{fig:framework}. The feature extractor includes a Patch Embedding layer, which map the input of a two-dimensional grid into non-overlapping multi-block sequences. These sequence data are then fed into four Transformer blocks configured with different scales, each generating feature maps with different resolutions. The scales of these feature maps are $1/4, 1/8, 1/16, and 1/32$ of the original input dimensions $(H, W)$, thereby achieving a multi-scale understanding of the input data.


In the feature fusion stage, corresponding to the multi-scale features, thera are four multimodal fusion modules. Features of the same scale under each modality are input into the same feature fusion module, thereby obtaining the fused features of four scales. In the Decoder stage, the multi-scale features will first pass through an MLP layer to obtain multi-scale features of the same channel, then be sampled to a uniform size $(H/4, W/4)$, and after channel concatenation and the next MLP layer, the multi-scale features are fused and used for predictive classification results.
By selecting a sufficiently powerful feature extractor for multi-scale feature extraction, the entire network can extract features from multiple modalities with fewer parameters. Meanwhile, our multimodal feature fusion module can effectively reorganize features, mining redundant features. At the same time, by fusing features from multiple scales, the final prediction results can integrate low-level spatial details with the semantic information of deep features. The subsection~\ref{sec:ff} will detail the entire multimodal feature fusion process.

\subsection{Multimodal Feature Fusion Module}
\label{sec:ff}
\begin{figure}[ht]
    \centering
    \includegraphics[width=0.8\linewidth]{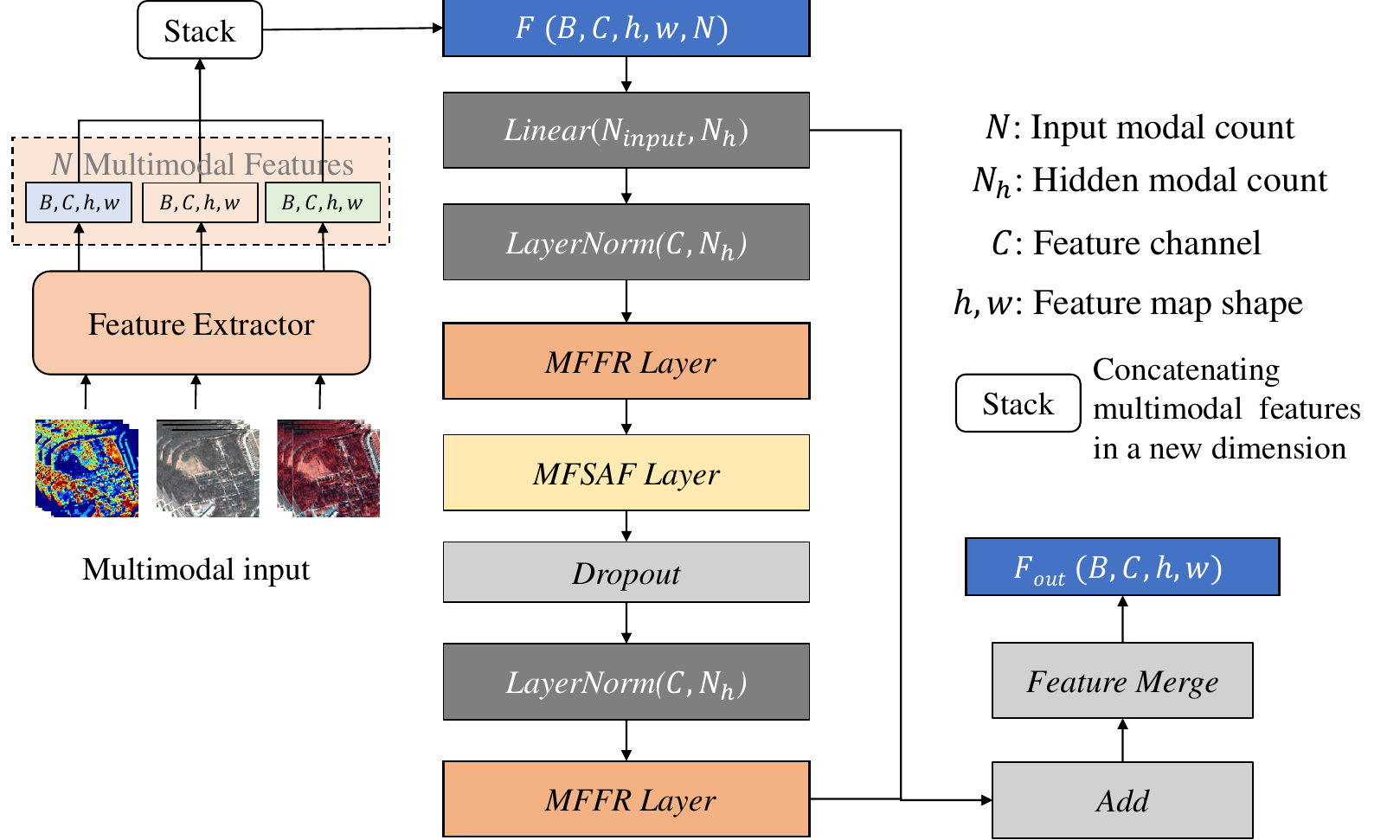}
    \caption{Multimodal feature fusion  and merge module.}
    \label{fig:fusion_module}
\end{figure}

Traditional cross-modal feature fusion methods mainly rely on pixel-wise summation and cascading operations, manually designing internal structures for the features of two modalities. This leads to limitations of these fusion methods when dealing with various types of data. Here, we manipulate multimodal features from a tensor perspective, achieving reorganization and fusion of multimodal features, allowing our method to handle inputs from multiple modalities. The fusion module we propose is based on MLP and Self-Attention, and the overall fusion process is depicted in Figure \ref{fig:fusion_module}. 

The multi-scale features extracted by the feature extractor are denoted as $F_m^i$, where $m$ represents the modality, and $i$ represents the layer number. Here, the value range of $i$ is $\{1,2,3,4\}$. For simplicity, $i$ will be omitted in the following text. In the fusion module, we stack the modal features of the same scale to form a five-dimensional multimodal feature tensor $F \in R^{B\times C\times h \times w \times N}$, where $C$ represents the number of channels, $h$ and $w$ are the width and height of the feature map, respectively, and $N$ is the number of modalities input. We propose a Multimodal Feature Fusion Reconstruction Layer (MFFR Layer) and a Multimodal Feature Self-Attention Fusion Layer (MFSAF Layer) to enhance fusion effects. The fusion process is described by Eq.~\ref{eq:fusion}.
\begin{equation}\label{eq:fusion}
\begin{aligned}
F_{r} &= LN(Linear(F)) \\
    F_{f} &= LN(Dropout(MFSAF(MFFR(F_{r})))) \\
    F_{res} &=F_{r} + MFFR(F_{f}) \\
    F_{out} &= FM(F_{res})
\end{aligned}
\end{equation}

In the Eq.~\ref{eq:fusion}, $LN$ stands for the Layer Normalization layer. After passing through the linear reproject module, features are expanded from $N$ modalities to $N_{h}$ implicit modalities, resulting in $F_r$.  The $F_r$ is then fed into the MFFR Layer ($MFFR$) and MFSAF Layer ($MFSAF$) for mixing of multimodal features. Subsequently, this feature is input into the MFFR Layer for reconstruction of multimodal features, thereby obtaining the feature $F_{f}$. Following that, the feature goes through Feature Merge ($FM$) operation , resulting in the fused feature $F_{out} \in R^{B\times C \times h \times w}$.

This step aims to increase the redundancy of modality information by extending it into a higher-dimensional space. In our fusion module, the normalization for different modalities is conducted separately because the mean and variance of each modality can to some extent represent the characteristics of the features of that modality. Therefore, to maintain their differences, we use separate normalization. Based on previous experience, in different layers, $N_{h}$ is set as $n, n+4, n+8, n+16$, respectively, where $n$ is a hyperparameter, the specific value of which is determined in the hyperparameter experiments.

The detailed structures of the MFFR Layer, the MFSAF Layer and $FM$ are introduced in subsections \ref{sec:mlplayer}, \ref{sec:attenlayer} and \ref{sec:maxop} respectively.

\begin{figure}[ht]
 
    \centering
    \subfloat[MFFR Layer]{\label{fig:mlpf}
 \includegraphics[width=.25\columnwidth]{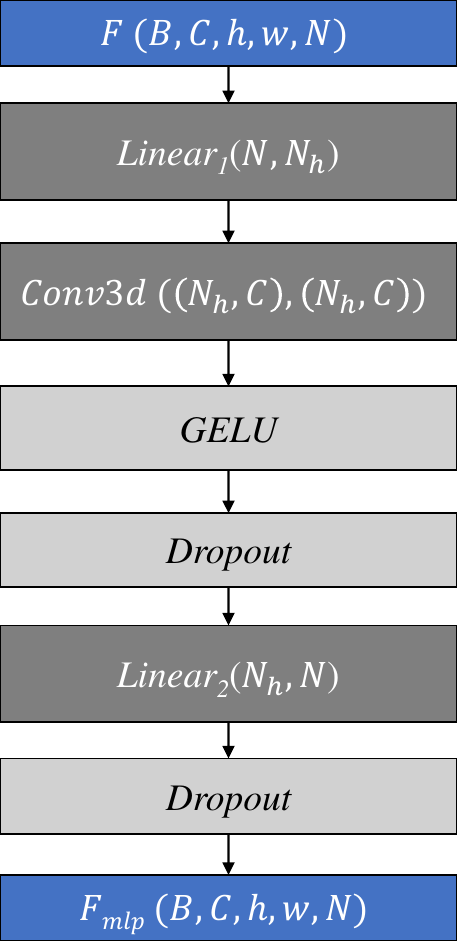}}\hspace{15pt}
 \subfloat[MFSAF Layer]{\label{fig:attf}
 \includegraphics[width=.45\columnwidth]{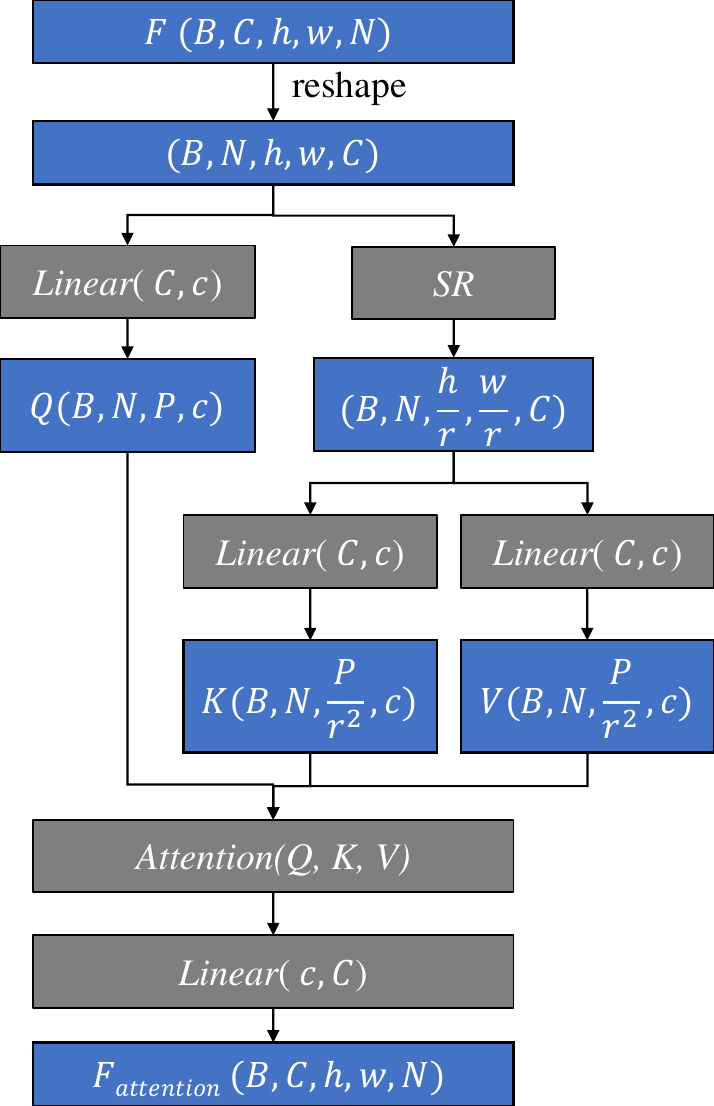}}\hspace{5pt}
 
    \caption{\label{fig:mlpatten}The structures of MFFR Layer and MFSAF Layer}
   
\end{figure}
\subsubsection{Multimodal Feature Fusion Reconstruction Layer}
\label{sec:mlplayer}


The structure of the \textbf{M}ultimodal \textbf{F}eature \textbf{F}usion \textbf{R}econstruction Layer (MFFR Layer) is shown in Figure \ref{fig:mlpf}. After a linear projection, the modal dimension of $F$ is expanded from $N$ to $N_{h}$. Then, we use a Group Conv3d module to calculate the local correlation between modalities. It is worth noting that here we specify the number of groups as the number of channels, allowing each convolutional kernel to focus on the features of the same channel under each modality separately. Finally, a linear layer is used to further fuse the features of each modality and restore them to the input quantity. The specific formula is shown in Eq.~\ref{eq:mlp}.

\begin{equation}\label{eq:mlp}
\begin{aligned}
    F_{1} & = Linear_1(F) \\
    F_{2} & = R_{inv}(Conv3d(R(F_{1}))) \\
    F_{mlp} & = Dropout(Linear_2(F_{2}))
\end{aligned}
\end{equation}
In Eq.~\ref{eq:mlp}, $F_1$, $F_2$ represent intermediate variables. $Linear_1$ and $Linear_2$ denote two linear layers. $R$ stands for the rearrange function, which transforms features of shape $(B, C, h, w, N)$ into a shape of $(B, N, C, h, w)$, matching the input format of Conv3d. $R_{inv}$ denotes the process of reversing this operation. The term $F_{mlp}$ represents the fused feature after MFFR Layer.

The MFFR Layer achieves the fusion and reconstruction among multimodal features through transformations in the modal dimension. $F_1$ represents its implicit expression, with each convolution kernel in the 3D convolution module responsible for an independent channel, achieving fusion in the latent space. The Linear layer for output represents the reconstruction process.

\subsubsection{Multimodal Feature Self-Attention Fusion Layer}
\label{sec:attenlayer}

The \textbf{M}ultimodal \textbf{F}eature \textbf{S}elf-\textbf{A}ttention \textbf{F}usion Layer (MFSAF Layer) is used for the cross-fusion of multimodal features.  Its structure is shown in Figure \ref{fig:attf}. The multimodal features $F$ go through three Linear projection modules to obtain three features: $Q$, $K$, $V$. Inspired by \cite{xie2021segformer}, to perform efficient computation, the size of $K$ and $V$ is resized by SR module. Specifically, we use a two-dimensional convolution with both kernel size and stride of $m \times m$ to reduce the size of $K$ and $V$ to $1/m^2$ of their original size, as shown in Eq.~\ref{eq:sr}.

\begin{equation}\label{eq:sr}
    SR(X) = Conv2d(kernel=m\times m, stride=m\times m)(X)
\end{equation}
We applied the modality fusion module at four different scales. Based on previous experience \cite{xie2021segformer}, we set the down-sampling stride $m$ at four scales to $\{16, 8, 4, 1\}$, respectively. 
In practical computations, it is also necessary to reshape the features to meet the input formats of various computational functions. For simplicity and to highlight the core process, the reshape step is omitted here. Subsequently, we treat the modal dimension $N$ as a head, so that for each modality, the following \textit{Attention} formula~\ref{eq:aten} applies:
\begin{equation}\label{eq:aten}
\begin{aligned}
Attention(Q, K, V)= Concate_{k=1}^{N} \{ softmax( \frac{Q_k\times K_k^T}{\sqrt{d_k}} ) \times V_k \}
\end{aligned}
\end{equation}
In the formula~\ref{eq:aten}, $k$ represents the $k-th$ modality. The same operation is performed for each modality, and they are recombined along the modal dimension to get the attention output $F_{attention}$.


\subsubsection{Multimodal Feature Merge Operations}
\label{sec:maxop}

After the feature fusion module, the shape of $F_{res}$  in Eq.~\ref{eq:fusion} is $(B,C,h,w,N)$. The operation of merging N into 1 is referred to as Multimodal Feature Merge Operations. There are generally three methods: $max$, $mean$, and $linear$, as shown in Eq.~\ref{eq:maxops}. 

\begin{equation}
\label{eq:maxops}
\begin{aligned}
max(F_{res}) & = max_{k = 1}^{N}F_{res}^{k} \\
mean(F_{res}) & = \frac{\sum_{k=1}^{N}F_{res}^{k}}{N} \\
linear(F_{res}) & = \frac{\sum_{k=1}^{N}w_kF_{res}^{k}}{N} + bias
\end{aligned}
\end{equation}
In the Eq.~\ref{eq:maxops}, $F_{res}^{k}$ is the $k$-th modal feature with shape of $(B, C, h, w)$. We empirically consider the $max$ to be the best because its advantage is that it is insensitive to the positions of modalities—this means that no matter which modality responds to the classification result, it will not significantly affect the value of most merge operation outputs. In the experiments, we test and discuss all three methods.

\subsection{MLP Decoder}
 


\begin{figure}
    \centering
    \includegraphics[width=0.75\linewidth]{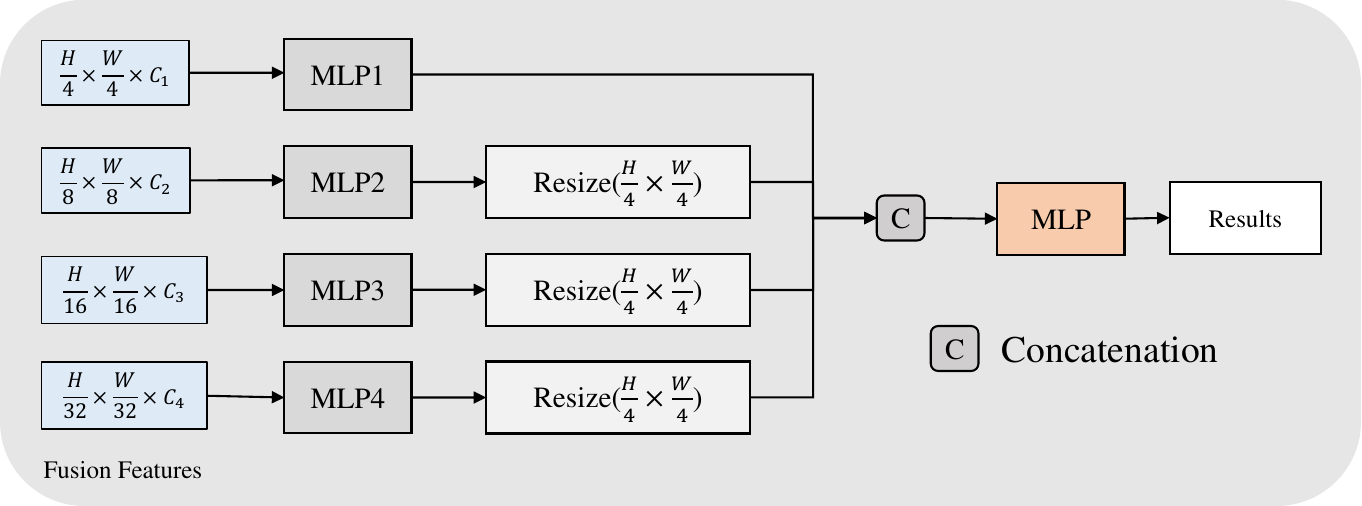}
    \caption{The MLP decoder structure.}
    \label{fig:mlpdecoder}
\end{figure}

The decoder structure in the network is shown in Figure~\ref{fig:mlpdecoder}. After multi-scale fused features pass through multiple MLP layers, they are sampled to the same size $(H/4, W/4)$, and then concatenated together to form multi-scale features. Subsequently, a simple MLP layer is used to obtain the classification results.

\subsection{Data Preprocessing}
\label{sec:dsmpre}

For multispectral data, we employ pseudo-color synthesis to decompose it into several three-band images. Specifically, for the R, G, B, and Nir bands, it is decomposed into RGB and NirRG two modalities. For single-band data similar to DSM, a colormap-based transformation method is applied to translate it into three-band image. This step is inspired by the manual interpretation process. In the manual image interpretation process, researchers typically utilize single-band pseudo-color to transform the color space to highlight elevation changes. Since the neural network aims to simulate the manual image interpretation process~\cite{zhang_learning_2017,feng2020deep}, the transformation of the color space also affects it. Many scholars have used transfer learning~\cite{zhen2020deeptransfer,feng2020deep} methods to complete this process, but this leads to a significant increase in training time and a larger amount of computation overall~\cite{feng2020deep}. Therefore, we opt for a colormap method that does not require additional parameters to complete this process.
\subsubsection{Normalization}

For the multi-spectral image, the mean and variance of each band across the entire dataset are firstly calculated, which are then used for the normalization of the multi-spectral data. As for the DSM, it is initially converted into RGB format, following which the normalization is carried out using the mean and variance from ImageNet.

During the process of converting DSM to RGB, since DSM represents the elevation of ground cover, the  distribution of it is different from that of others. Here, we perform a simple clip on DSM using the percentile clip method shown in Eq.~\ref{eq:clip}:

\begin{align} \label{eq:clip}
Clip(X) & = \left\{\begin{matrix} 
  p_{\alpha1}(X), X \leq p_{\alpha1}(X) \\
  p_{\alpha2}(X), X \geq p_{\alpha2}(X) \\
 X, others
\end{matrix}\right. 
\end{align}
where, $p_{\alpha}(X)$ represents the $\alpha$ percentile of $X$, with $X$ being the DSM. Specifically, in the experiments, $\alpha_1=5$ and $\alpha_2=95$. For normalization, we adapt the local min-max method to scale its data range to 0-1, as shown in Eq.~\ref{eq:minmax}.
\begin{equation}\label{eq:minmax}
     \begin{aligned}
X_c & = Clip(X) \\
X_n & = \frac{(X_c - min(X_c))}{(max(X_c) - min(X_c)}
\end{aligned}
\end{equation}

 Subsequently, we apply the jet colormap to convert the grayscale values into rgb space. The comparision among different colormap is in the~\ref{sec:colormaps}.





\section{Experiments and Analysis}
\label{sec:exps}
\subsection{Datasets}

To verify the effectiveness of the proposed method, we conduct experiments on three datasets: US3D, ISPRS Potsdam, and ISPRS Vaihingen. Among these, ablation studies and hyperparameter experiments are carried out on the US3D dataset, and comparative experiments with state-of-the-art (SOTA) methods are performed across all datasets. Below, each of the three datasets is introduced in detail.

\subsubsection{US3D Dataset}

US3D~\cite{bosch2019semantic} is a remote sensing semantic stereo dataset with images from Wordview-3 multi-view imagery, containing 26,709 training and validation images and 14,529 test images, each of size $2048 \times 2048 $ pixels. The primary task is stereo image matching and disparity generation, with DSM data and a five-class (Ground, Foliage, Building,  Water and Bridge) surface classification map provided, generated according to the Homeland Security Infrastructure Program (HSIP) and manually edited where necessary. The imagery includes seven bands, but in the experiments, only the red (R), green (G), blue (B), and near-infrared (Nir) bands are used, inputted as RGB and NirRG band combinations. For the DSM data, since the dataset does not provide normalized DSM (nDSM) data, we normalize and render it following the method described in~\ref{sec:dsmpre}. During training, to improve efficiency, patches of $512 \times 512$ pixels are randomly selected from the images as input. For prediction, the process is carried out with a stride of 300 pixels and a patch size of $512 \times 512$ pixels.

\subsubsection{ISPRS Potsdam Dataset}
The ISPRS Potsdam dataset consists of 38 true orthophoto (TOP) images and corresponding nDSM images. Both the TOP and nDSM have a spatial resolution of $5 cm$, with pixels of $6000 \times 6000$. The TOP includes Red (R), Green (G), Blue (B), and Near-Infrared (Nir) channels. The ground-truth contains five main land cover categories: impervious surfaces, buildings, low vegetation, trees, and cars. According to the data provider, the training set is composed of twenty-four patches, with the remaining patches used as the test set. We utilized 14 images for testing,  1 for validation, and the remaining 22 images for training. During the experiments, for the TOP we use two band combinations for input: RGB and NirRG. For the nDSM data, we use the RGB output after jet colormap as the experimental input. All data are cropped to a size of $512 \times 512$ pixels.
 
\subsubsection{ISPRS Vaihingen Dataset}

In the ISPRS Vaihingen dataset, each patch is cut from a considerably large TOP associated with the town of Vaihingen. The sizes of the patches vary from each other, approximately $2000 \times 2500$ pixels, with a spatial resolution of $9 cm$. Each patch contains only three bands: Near-Infrared (Nir), Red (R), and Green (G). Additionally, a corresponding nDSM is provided for each image block. The categories are the same as those in the Potsdam dataset. According to the data provider, 16 of all patches are used as a training set, with the remaining patches used for testing. For the nDSM data, we use the RGB output after jet colormap as the experimental input. All data are cropped to a size of $512 \times 512$ pixels. 

\subsection{Accuracy Assessment}

The quality and performance of deep learning models are usually evaluated by analyzing their accuracy on test data. In this paper, we use $mF1$ score, overall accuracy ($OA$), intersection over union ($IoU$) for each category, and mean IoU ($mIoU$) to assess the performance of models. Furthermore, we also calculate the number of parameters for each model. To compute the aforementioned accuracy metrics, we first introduce the multi-class confusion matrix $M$~\ref{eq:m}.
\begin{equation}\label{eq:m}
    M_{ij}= \sum{ y=i \cap \hat{y}=j }    
\end{equation}
Typically,  $OA$ is used to evaluate segmentation performance. This metric is the ratio of the number of pixels successfully predicted in all patches in the test dataset to the total number of pixels, calculated as Eq.~\ref{eq:oa}.
\begin{equation}\label{eq:oa}
    OA = \sum{M_{ii}}/\sum M
\end{equation}
The calculation of $mF1$ is slightly complex, as shown in Eq.~\ref{eq:mf1}. 
\begin{equation}\label{eq:mf1}
    \begin{aligned}
F1_i & = \frac{2TP_i}{2TP_i+FN_i+FP_i} \\
TP_i & = M_{ii} \\
FN_i &= \sum_{j \neq i}M_{ij} \\
FP_i & = \sum_{j \neq i}M_{ji} \\
mF1 &= \sum{F1_i} / C
\end{aligned}
\end{equation}
The $C$ in Eq.~\ref{eq:mf1} denotes the count of classes. $IoU$ is a metric for measuring the degree of overlap between the true image and the prediction result. The calculation formulas for the $IoU$ of each category and the $mIoU$ are as Eq.~\ref{eq:iou} : 

\begin{equation}\label{eq:iou}
    \begin{aligned}
IoU_i & = \frac{M_{ii}}{ \sum_{j \neq i } M_{ij} +   \sum_{j \neq i } M_{ji}  } 
\\
mIoU & =  \sum{IoU_i} / C
\end{aligned}
\end{equation}

\subsection{Implementation details}

In this work, we implement all network frameworks using MMSegmentation~\cite{mmseg2020} and PyTorch 2.1.0. All networks are trained in parallel on four Tesla V100 GPUs. The batch size is set to 4 and the optimizer used is AdamW~\cite{loshchilov2017decoupled}. Weight decay for layers other than normalization layers is set to 0.01. The initial learning rate is set to $6\times 10^{-5}$. A warm-up strategy is adopted as the change schedule in learning rate, where it linearly increased from $1\times 10^{-6}$ to $6\times 10^{-5}$ over $0 - 1500$ steps. After that, it decayed to 0 following the formula $(1-\frac{iter-t_0}{max-t_0-iter})^{p}$ , where $t_0$ is set to 1500, $p = 0.9$. The $max$ is $8\times 10 ^ 4$ for US3D dataset, and for Potsdam and Vaihingen dataset, it is  $4\times 10 ^ 4$. To mitigate the phenomenon of overfitting, data augmentation is introduced during training, including random horizontal flips, vertical flips, random scaling, and random rotations, etc.

The input data is fixed at a width and height of $512\times 512$ pixels. In the experiments, all involved backbones adopted pretrained weights from ImageNet. For those without pretrained weights, the Kaiming Initialization~\cite{he2015delving} method is used for initialization. Regarding the loss function, all experiments use the same cross-entropy loss function without any modifications.

\subsection{Ablation study}

This section investigates the effectiveness of the network fusion module. Ablation studies are carried out by independently removing and altering each key component of the network. The specific experimental setups are as follows: 
\begin{itemize}
    \item baseline-RGB: The Segformer model that takes only the RGB band as input. 
    \item baseline-NirRG: The Segformer model that takes only the NirRG band as input. 

    \item cat-only: RGB, NirRG and DSM as inputs but employs a simple band concatenation method at the data fusion stage to demonstrate that multimodal data significantly boosts  accuracy compared to single-modality input, where the DSM data is in rgb format.
    \item MFFR: only the MFFR Layer is used, removing the MFSAF Layer depicted in Figure \ref{fig:framework}.
    \item MFSAF: Only the MFSAF Layer is used, removing the MFFR Layer depicted in Figure \ref{fig:framework}.
        \item RGB+DSM:  RGB and DSM as inputs, other settings are the same as those in LMFNet.
    \item NirRG+DSM: NirRG and DSM as inputs, other settings are the same as those in LMFNet.

    \item LMFNet (ours): The LMFNet we proposed, where the fusion module adopts the structure shown in Figure~\ref{fig:framework}, with the 
RGB, NirRG and colormaped-DSM as inputs.

\end{itemize}
The quantitative evaluation results of the experiment are presented in Table \ref{tab:ab}.

\begin{table}
 \caption{Ablation quantitative results on the US3D dataset. The blod is the best.}
\begin{adjustbox}{width=1\columnwidth,center}

    \centering
    
    \begin{tabular}{ccccccccc} 
     \hline
     \multirow{2}{*}{Exps.} & \multicolumn{5}{c}{$IoU$ (\%)}   &  \multirow{2}{*}{$mF1$ (\%)}&  \multirow{2}{*}{$mIoU$ (\%)}&  \multirow{2}{*}{$OA$ (\%)}\\
     \cline{2-6}
          &  Ground&  Foliage&  Building&  Water&  Bridge&   & &
          \\ 
         \hline
         baseline-RGB&  84.37&  62.77&  70.47&  81.72&  59.67&   83.20&71.80&87.24
\\
 baseline-NirRG& 84.69& 63.01& 70.70& 81.92& 59.81& 83.27& 72.03&87.85
 \\ 
         cat-only&  91.92&  78.11&  79.48&  83.01&  85.05&   90.94&83.51&92.99
\\ 
         MFFR&  92.46 &  79.05&  80.15&  84.86&  85.43&   91.46&84.39&93.36 
         \\
 MFSAF& 92.23& 78.81& 80.06& 83.08& 84.90&  91.12&83.82&93.22
\\
 RGB$+$DSM  &92.53 &79.46 &80.73 &85.48 &86.94 &91.47 &85.03 &93.40
\\
NirRG$+$DSM&92.58&79.72&80.83&85.51&87.02&91.62&85.13&93.50
\\
 LMFNet (ours) & \textbf{92.61}&\textbf{79.52}& \textbf{80.83}&\textbf{85.61}&\textbf{86.90}&  \textbf{91.88}&\textbf{85.09}&\textbf{93.53}
\\
\hline
    \end{tabular}

   \end{adjustbox}
    \label{tab:ab}
\end{table}

\subsubsection{Multimodal Fusion Module}

In terms of utilizing multimodal data, we compare four sets of experiments, including: baseline-RGB (RGB input), baseline-NirRG (NirRG input), cat-only (three inputs), and LMFNet (three inputs and fusion module). It can be discerned from Table \ref{tab:ab} that, compared to single-modal inputs, multimodal data can significantly improve result accuracy. Even if only a simple concatenation fusion is carried out in the Decoder phase, it can increase the $mIoU$ by $13\%$ and the $OA$ by $6\%$. This indicates that elevation information greatly aids in land cover classification. Moreover, after adopting our fusion method, the $mIoU$ has increased by another $1.5$ percentage points. 

Furthermore, observing the classification performance of individual categories, the accuracy of each category has significantly improved. Among them, the accuracy of bridges has increased the most, with an $IoU$ increase of $25.75\%$. The texture of bridges is extremely similar to that of roads, but there is a significant difference in elevation information between the two, hence the accuracy greatly increased after adding elevation data. Trees have increased by approximately $16$ percentage, and buildings have also improved by about $10$ percentage due to the addition of elevation information. Water have a unique texture, so the accuracy improvement for water is the smallest, but there is also an increase of $3$ percentage.

The inclusion of the DSM modality effectively improves the accuracy of land cover classification, even through the simplest method of channel concatenation. Accuracy can be further enhanced after adopting our fusion module. 

\subsubsection{Subfusion Layer Results}

To investigate the role our fusion module plays and why it can improve classification accuracy, we separately remove the MFSAF Layer and the MFFR Layer for comparison. The experiments compared are: MFSAF, MFFR, LMFNet (ours). As it can be seen from Table~\ref{tab:ab}, whether adding the MFFR Layer or the MFSAF Layer, both improve accuracy. Comparatively, the improvement from the MFFR Layer is better, with the effect of the MFSAF Layer being slightly inferior.


\subsubsection{Different Modal Combinations}

We compare the results of combining RGB with DSM data against combinations with NirRG. The experimental findings suggest that the NirRG combination is more advantageous for the classification of ground objects in remote sensing. The network with NirRG $+$ DSM as input had an accuracy improvement of 0.3 percentage points in the Foliage category. The NirRG combination includes the near-infrared band, which offers a significant benefit for vegetation since vegetation reflects much more in the near-infrared band than in the visible light wavelengths. This helps the model more easily differentiate between vegetated and non-vegetated areas.

\subsection{Hyperparameter Experiment}

In our method, multiple hyperparameters are involved, such as the dimensionality reduction method and the number of hidden modalities $N_h$ in the fusion module. In this section, we conduct experiments on different hyperparameter schemes and selected the optimal hyperparameters through quantitative evaluation.

\subsubsection{The Number of Hidden Modalities }

\begin{figure}[ht]
    \centering
    \includegraphics[width=1\linewidth]{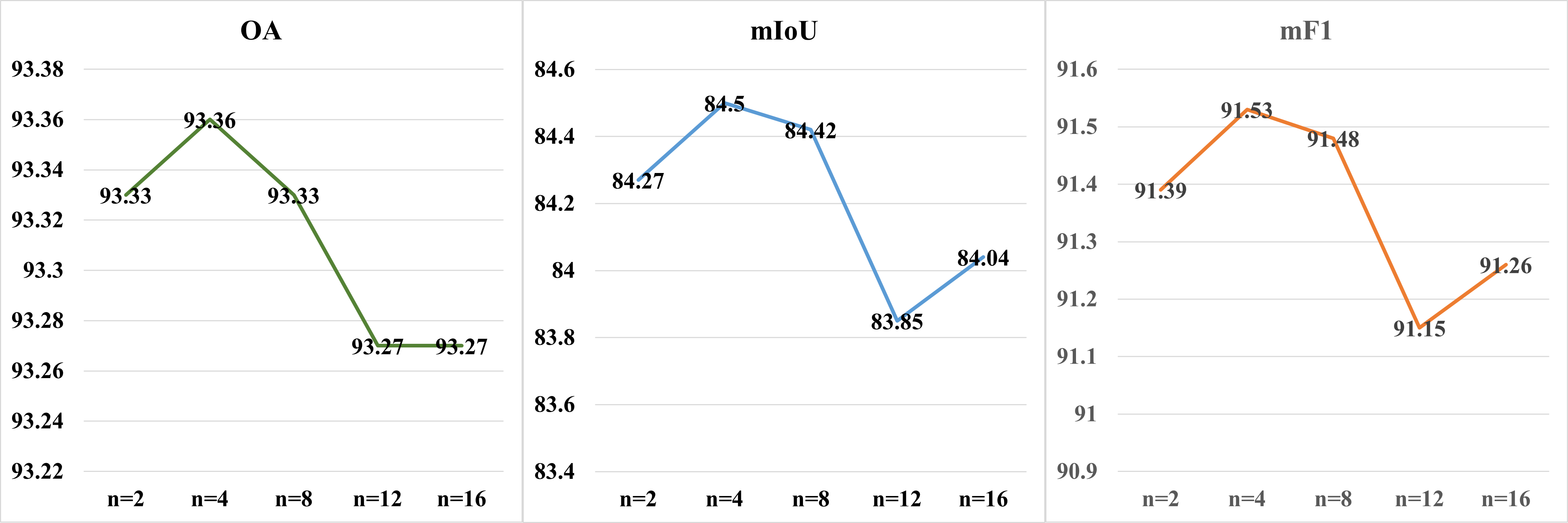}
    \caption{The relationship between the number of hidden modalities and accuracy. The graph shows the $OA$, $mIoU$, and $mF1$ for different values of $n$.}
    \label{fig:ntest}
\end{figure}

In Section \ref{sec:ff} of this paper, it is stated that we set the $N_h$ at different scales to $n, n+4, n+8, n+16$. To find the most suitable parameters, we test scenarios with $n=2, 4, 8, 12, 16$. Due to the time cost of the experiments, they are conducted only on the US3D dataset. The quantitative accuracy evaluation is illustrated in Figure~\ref{fig:ntest}. It is found that the best performance occurs when $n=4$.

\subsubsection{Multimodal Feature Merge Opertation}

In the multimodal feature merging stage, we compare three merge opertations: $max$, $mean$, and $linear$. The quantitative results are shown in Table \ref{tab:meanmax}.  LMFMet-max is the method employed in this study with $max$ operation. LMFNet-mean uses the $mean$ method for merge opertation, and LMFNet-linear utilizes the $linear$ method. 

\begin{table}
\caption{Evaluation metrics of different feature merge opertation in our methods in US3D dataset}
    \centering
\begin{adjustbox}{width=1\columnwidth,center}
\label{tab:meanmax}
    \begin{tabular}{ccccccccc}
    \hline
        \multirow{2}{*}{Methods} & \multicolumn{5}{c}{$IoU$(\%)}   &  \multirow{2}{*}{$mF1$(\%)} &  \multirow{2}{*}{$mIoU$(\%)} &  \multirow{2}{*}{$OA$(\%)}
     \\ 
     \cline{2-6}
          &  Ground&  Foliage&  Building&  Water&  Bridge&   & &\\ 
         
      \hline
         LMFNet-mean&  92.45&  78.98&  80.33&  84.22&  85.63&  91.42&  84.32& 93.37
\\
         LMFNet-linear&  92.50&  79.29&  80.31&  83.34&  86.09&  91.41&  84.31& 93.40
\\
         LMFNet-max&  \textbf{92.61}&  \textbf{79.52}&  \textbf{80.83}&  \textbf{85.61}&  \textbf{86.90}&  \textbf{91.88}&  \textbf{85.09}& \textbf{93.53}
\\
\hline
    \end{tabular}
    \end{adjustbox}
    
\end{table}


It is found that LMFNet-max has the best accuracy, with the scores of $mean$ and $linear$ methods being similar. In practice, the $mean$ method can be equivalent to a linear method with a weight of $1/m$, where $m$ is the number of modalities and $bias=0$, hence their similar performance. Therefore, using the maximum value for dimensional reduction can effectively improve the model's utilization of multimodal input, thereby enhancing the classification accuracy. Detailed analysis and discussion can be found in Section~\ref{sec:discusiion}.

\subsection{Comparison with SOTA methods}

Currently, for the task of semantic segmentation of VHR data, there are many excellent models, and we implement those for comparison with our method. We select the US3D dataset, ISPRS Potsdam dataset, and ISPRS Vaihingen dataset to evaluate the performance of our method and others. For the single-modal model, we select models including PSPNet~\cite{zhao2017pyramid}, DeeplabV3+~\cite{chen2018encoder}, HRNet~\cite{wang2020deep}, UNetFormer~\cite{wang_unetformer_2022} and Segformer~\cite{xie2021segformer}, etc. For multimodal model, we chose FuseNet~\cite{lai_fusenet_2017}, vFuseNet~\cite{audebert2018beyond}, SA-Gate~\cite{chen2020bi}, CMGFNet~\cite{hosseinpour_cmgfnet_2022}  and CMX~\cite{zhang2023cmx}for comparison. To ensure a fair comparison, all multimodality methods used colormap DSM. Moreover, for the selection of bands for multi-spectral data, we referred to the results in Table~\ref{tab:ab} and select NirRG as the input for multi-spectral data. As for our LMFNet, since it can accept both bi-modal and tri-modal inputs, we named the model for bi-modal input LMFNet-2, and for tri-modal input, it is named LMFNet-3. 

\subsubsection{Results on the US3D Dataset}

\begin{figure}
    \centering
    \includegraphics[width=0.9\linewidth]{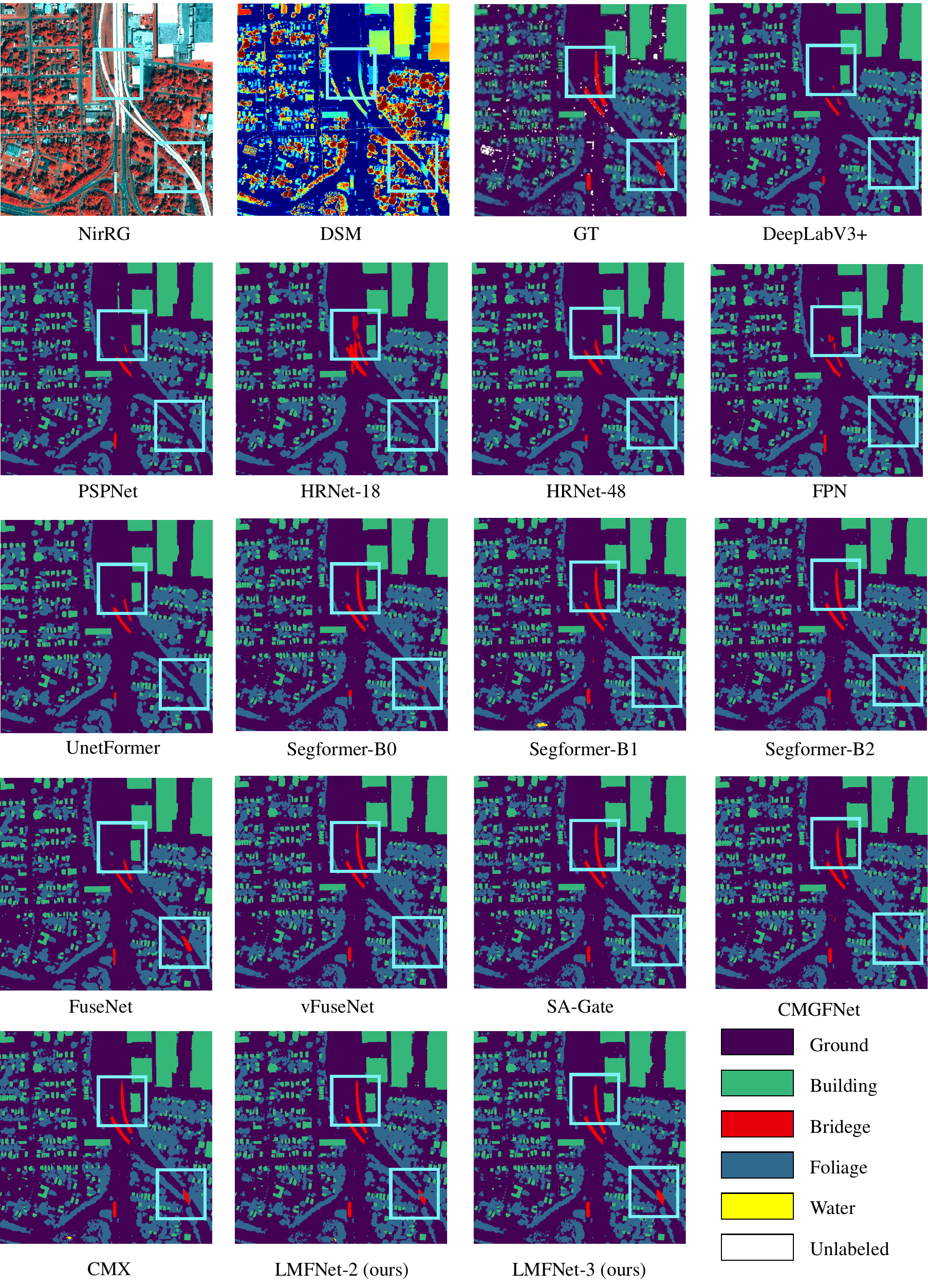}
    \caption{Qualitative comparison on the US3D dataset between ours and other SOTA methods. The light blue boxes highlight the areas of focus. DSM is represented by jet colormap. }
    \label{fig:us3ddemo}
\end{figure}

The quantitative accuracy evaluation results and on the US3D dataset are shown in Table \ref{tab:us3d}. The parameters of the model are also counted. We calculate the $IoU$ for each category as well as the  $mF1$, $mIoU$, and  $OA$.  The Modals column in the Table \ref{tab:us3d} represents the input modalities. The Figure \ref{fig:us3ddemo} presents the visualized prediction results.
\begin{table}
\caption{The quantitative results of our method vs. others on the US3D dataset. The optimal results for the number of parameters on the single-modal and multimodal models are highlighted in bold respectively.}
\begin{adjustbox}{width=\columnwidth,center}
    \centering

\label{tab:us3d}
    \begin{tabular}{ccccccccccc}
    \hline
          \multirow{2}{*}{Modals} & \multirow{2}{*}{Methods} & \multicolumn{5}{c}{$IoU$ (\%)}    &  \multirow{2}{*}{$mF1$ (\%)}&  \multirow{2}{*}{$mIoU$ (\%)}&  \multirow{2}{*}{$OA$ (\%)} &\multirow{2}{*}{Params (M)}\\   \cline{3-7}
        &  &  Ground&  Foliage&  Building&  Water&  Bridge&   &  & &\\  
      
      \hline
      \multirow{9}{*}{NirRG }&PSPNet&  84.89&  63.10&  72.64&  80.59&  64.39&  84.19& 73.12&87.70
  &
48.97\\
  &DeepLabV3+& 85.45& 64.19& 73.4& 82.15& 66.91& 85.08& 74.42&88.13
  &43.58\\
          &HRNet-18&  79.91&  57.38&  57.99&  74.48&  31.9&  73.78&  60.33&83.49
  &
9.63\\
  &HRNet-48& 84.61& 62.49& 71.8& 81.11& 63.29& 83.85& 72.66&87.45
  &65.85\\
  &UNetFormer& 85.94& 65.12& 74.27& 82.04& 66.5& 85.31& 74.77&88.51
  &
24.20\\
  &FPN& 84.44& 62.56& 71.16& 80.75& 63.65& 83.76& 72.51&87.30
  &28.49\\
  &Segformer-B0& 84.37& 62.77& 70.47& 81.72& 59.67& 83.20& 71.80&87.24
  &
3.71\\
  &Segformer-B1& 85.32& 64.37& 72.57& 82.97& 65.70& 84.90& 74.19&88.02
  &13.68\\
  &Segformer-B2& 85.76& 64.98& 73.52& 83.48& 67.93& 85.55& 75.14&88.37
  &
24.72\\
 \hline  
 
 \multirow{6}{*}{\begin{tabular}[c]{@{}l@{}}NirRG\\ DSM\end{tabular}}&FuseNet& 91.96& 78.1& 79.8& 83.6& 84.73& 91.02& 83.64&93.03
 &42.08\\
 
 &vFuseNet& 91.97& 78.16& 79.77& 82.38& 85.36& 90.95& 83.53&93.01
 &
44.17\\
  &SA-Gate& 91.98& 78.04& 79.93& 83.27& 84.90& 91.01& 83.62&93.04
 &51.35\\
    &CMGFNet& 92.41& 79.11& 80.24& 82.85& 85.45& 91.24& 84.01&93.33
 &
123.63\\
        &CMX& 92.28& 78.91& 80.66& 85.06& 86.23& 91.61& 84.63&93.31
 &11.19\\
        &LMFNet-2 (ours)& 92.43& 79.09& 80.26& 84.99& 85.75& 91.53& 84.50&93.36 &
3.72\\
     \hline
 \begin{tabular}[c]{@{}l@{}}NirRG\\ RGB\\ DSM\end{tabular}&LMFNet-3 (ours)& \textbf{92.61}& \textbf{79.52}& \textbf{80.83}& \textbf{85.61}& \textbf{86.90}& \textbf{91.88}& \textbf{85.09}&\textbf{93.53}
 &4.22\\
 \hline
    \end{tabular}
    
    \end{adjustbox}
\end{table}

From the Table~\ref{tab:us3d}, it is clear that on the US3D dataset, multimodality methods generally outperform single modality approaches. Notably, the accuracy for bridges has seen significant improvement. Among the bi-modal-input models, the accuracy of CMX is slightly better than LMFNet-2, but with the NirRG, RGB and DSM as inputs, our method (LMFNet-3) achieved the best accuracy results.  Other networks are difficult to transform to accept trimodal inputs. Additionally, as we increase the number of modalities, the number of parameters in our network increases from $3.72M$ (LMFNet-2) to $4.22M$ (LMFNet-3), which is far less than that of other multimodal methods. Other multimodal methods face difficulties in directly increasing the number of modalities to achieve improvements in accuracy.

We conduct a qualitative comparison using images from the test set, as shown in Figure \ref{fig:us3ddemo}. In the figure, GT represents the ground truth. Comparing the recognition details of other methods (highlighted by the light blue boxes in the Figure~\ref{fig:us3ddemo}), single-modality methods are also capable of identifying parts of bridges, possibly through the assistance of shadows in the image. Therefore, in areas with substantial tree cover, the recognition of bridges is relatively poor. After incorporating DSM, except for FuseNet, the recognition accuracy of bridges has improved across the board. Additionally, in our results, all bridges are identified with commendable precision. 

\subsubsection{Results on the Potsdam Dataset}

The quantitative assessment on the Potsdam dataset is shown in Table~\ref{tab:potsdam}. It can be seen that our LMFNet-3 achieved the highest $mIoU$, $mF1$, and $OA$, which are 86.39\%, 92.36\%, and 92.72\%, respectively. Compared to the single-modal method (Segformer-B2), our $mIoU$ increased by 1.74\%. Compared to the bimodal method (CMX), it improved by 0.42\%. In all categories except for the car, our LMFNet-3 achieved the highest accuracy in impervious, building, low vegetation, and tree categories, which are 89.38\%, 95.31\%, 82.11\%, and 79.05\% respectively. Compared to the single-modal method (with Segformer-B2), the $IoU$ increased by 1.21\%, 0.53\%, 4.62\%, and 2.72\%, respectively. Low vegetation and tree categories have the most significant improvements. Compared to the bimodal method (CMX), the $IoU$ in each category increased by 0.54\%, 0.55\%, 0.17\%, and 0.99\%, respectively. 

\begin{table}[ht]
\caption{The quantitative results of our method vs. others on the Potsdam dataset.}
\begin{adjustbox}{width=\columnwidth,center}
    \centering

\label{tab:potsdam}
    \begin{tabular}{cccccccccc}
    \hline
     \multirow{2}{*}{Modals} & \multirow{2}{*}{Methods} & \multicolumn{5}{c}{$IoU$ (\%)}   &  \multirow{2}{*}{$mF1$ (\%)} &  \multirow{2}{*}{$mIoU$ (\%)} &\multirow{2}{*}{$OA$ (\%)}
       \\
       \cline{3-7}
        &  &  impervious&  building &  low vegetation&  tree&  car&   & & \\ 
          
      \hline
      \multirow{9}{*}{NirRG}&PSPNet&  87.09&  94.04&  75.61&  74.95&  83.21&  90.89& 82.98 &91.18
\\
  &DeepLabV3+& 87.15& 94.28& 75.93& 76.07& 84.90& 90.95& 83.67&91.39

\\
    &HRNet-18&  85.76&  91.74&  75.27&  74.29&  83.76 &  90.12&  84.02
&90.46
\\
  &HRNet-48& 88.01& 94.79& 77.12& 76.44& 84.75& 91.49& 84.22 &91.84
\\
  &UnetFromer& 88.04& 94.69& 77.15& 76.45& 84.42& 91.52& 85.18 &91.84
\\
  &FPN& 86.59& 93.41& 75.69& 74.45& 84.79& 90.54& 82.99&90.94
\\
  &Segformer-B0& 87.25& 93.94& 76.54& 75.17& 85.44& 90.95& 83.67&91.35
\\
  &Segformer-B1& 87.85& 94.54& 77.28& 76.12& 86.08& 91.37& 84.37&91.76
\\
  &Segformer-B2& 88.17& 94.78& 77.49& 76.33& 86.46& 91.53& 84.65&91.91
\\
 \hline  
 
 \multirow{6}{*}{ \begin{tabular}[c]{@{}l@{}}NirRG\\ DSM\end{tabular}}
 &FuseNet& 87.10& 93.54& 79.27& 76.91& 84.20 & 91.21& 84.20&91.48
\\
 
 &V-FuseNet& 87.91& 93.96& 79.60& 77.14& 83.21& 91.36& 84.36&91.77
\\
  &SA-Gate& 87.61& 94.92& 76.91& 75.42&86.56 & 91.09& 84.28&91.62
\\
    &CMGFNet& 88.67& 93.01& 80.73& 77.17& \textbf{86.83} & 91.54& 85.28 &91.97
\\
        &CMX& 88.84& 94.76& 81.94& 78.06& 86.27 & 92.15& 85.97&92.48
\\
        &LMFNet-2 (ours)& 89.39& 93.81& 81.26& 78.14&84.93 & 91.81& 85.51&92.20
\\
     \hline
 \begin{tabular}[c]{@{}l@{}}NirRG\\ RGB\\ DSM\end{tabular}&LMFNet-3 (ours)& \textbf{89.38}& \textbf{95.31}& \textbf{82.11}& \textbf{79.05}& 86.08& \textbf{92.36}& \textbf{86.39}&\textbf{92.72}
 \\
 \hline
    \end{tabular}
    
    \end{adjustbox}
\end{table}

\begin{figure}
    \centering
    \includegraphics[width=0.9\linewidth]{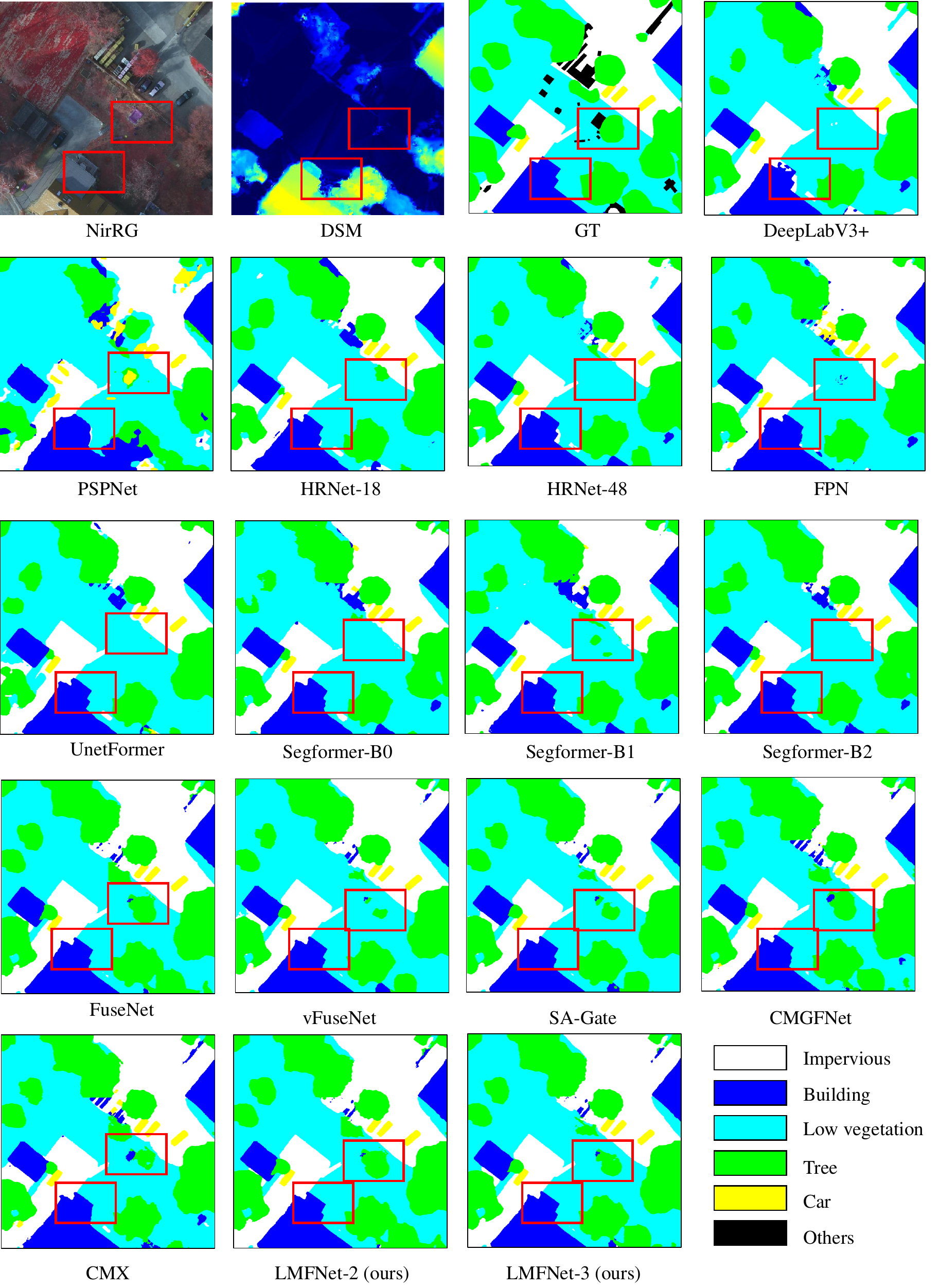}
    \caption{Qualitative comparison on the Potsdam dataset between ours and other SOTA methods. The red boxes in the figure mark the areas of focus. }
    \label{fig:potsdamdemo}
\end{figure}

We select samples from the test set for qualitative evaluation, and the results are shown in Figure~\ref{fig:potsdamdemo}. From the figure, it can be seen that due to some trees being too similar in color to grass, the recognition effect is poor in single-modality detection. With the DSM data as multimodal input, trees can be more easily recognized. Our method provides better visual effects for both building boundaries and trees.

\subsubsection{Results on the Vaihingen Dataset}

The quantitative results on the Vaihingen dataset are shown in Table \ref{tab:vai}. Our method achieves $IoU$ on impervious, building ,  low vegetation,  trees and  car  of 85.49\%, 91.46\%, 80.34\%, 82.85\%, and 72.33\%, respectively. Compared to the single-modal method (Segformer-B2), the improvements are 2.82\%, 1.93\%, 14.27\%, 7.13\%, and 2.86\% respectively. It is noticeable that the accuracy improvements for trees and low vegetation are the most significant. Since the Vaihingen dataset only includes NIR, R, G, lacking the B band, our method can only process data from two modalities, resulting in an overall accuracy slightly lower than CMX. However, our method still achieves the highest $IoU$, 91.46\%, for buildings. Moreover, our method (LMFNet-2) also has the smallest number of parameters (3.72M) among the multimodal methods.

\begin{table}[ht]
\caption{The quantitative results of our method vs. others on the Vaihingen dataset.}
\begin{adjustbox}{width=1\columnwidth,center}
    \centering

\label{tab:vai}
    \begin{tabular}{ccccccccccc}
    \hline
         \multirow{2}{*}{Modals} & \multirow{2}{*}{Methods} & \multicolumn{5}{c}{$IoU$(\%)}   &  \multirow{2}{*}{$mF1$(\%)} &  \multirow{2}{*}{$mIoU$(\%)} &  \multirow{2}{*}{$OA$(\%)} & \multirow{2}{*}{Params (M)} 
     \\
     \cline{3-7}
        &  &  impervious&  building &  low vegetation&  tree&  car&   &  & &\\ 
      \hline
      \multirow{9}{*}{NirRG}&PSPNet&  82.07&  89.14&  66.64&  76.08&  70.02&  86.63& 76.79&88.12 &48.97
\\
  &DeepLabV3+& 78.61& 84.08& 63.59& 73.90& 64.30& 84.08& 72.90&85.92 &43.58
\\
          &HRNet-18&  82.90&  88.24&  64.90&  75.67&  67.81&  86.03&  75.90&87.81 &9.63
\\
  &HRNet-48& 82.54& 89.42& 66.65& 75.04& 70.11& 86.60& 76.75&88.00 &65.85
\\
  &UnetFromer& 82.22& 89.54& 66.96& 75.98& 71.49& 86.93& 77.24&88.24 &24.20
\\ 
  &FPN& 81.15& 87.61& 65.74& 75.56& 64.23& 85.32& 74.86&87.46 & 28.49
\\
  &Segformer-B0& 81.34& 87.96& 65.79& 75.72& 67.05& 85.83& 75.57&87.60 & 3.71
\\
  &Segformer-B1& 82.56& 89.45& 66.83& 76.18& 69.60& 86.71& 76.92&88.27 & 13.68
\\
  &Segformer-B2& 82.67& 89.53& 66.07& 75.72& 69.47& 86.54& 76.69&88.13 & 24.72
\\
 \hline  
 
 \multirow{6}{*}{ \begin{tabular}[c]{@{}l@{}}NirRG\\ DSM\end{tabular}}
 &FuseNet& 81.81& 89.86& 73.83& 78.92& 68.60& 87.84& 78.60&88.87 & 42.08
\\
 
 &V-FuseNet& 82.88& 89.17& 78.47& 80.19& 68.07& 88.48& 79.56&89.62 & 44.17
\\
  &SA-Gate& 83.48& 89.22& 78.53& 81.83& 72.11& 89.47& 81.03&91.22 & 51.35
\\
    &CMGFNet& 85.08& 89.08& 80.87& 81.50& 71.29& 89.82& 81.56&90.89 & 123.63
\\
        &CMX& \textbf{87.30}& 90.21& \textbf{80.92}& \textbf{83.22}& \textbf{72.7}& \textbf{90.70}& \textbf{82.87}&\textbf{92.21} & 11.19
\\
        &LMFNet-2 (ours)& 85.49& \textbf{91.46}& 80.34& 82.85& 72.33& 90.44& 82.49&92.02 & 3.72
        \\
    \hline
    \end{tabular}
    
    \end{adjustbox}
\end{table}

\begin{figure}
    \centering
    \includegraphics[width=0.9\linewidth]{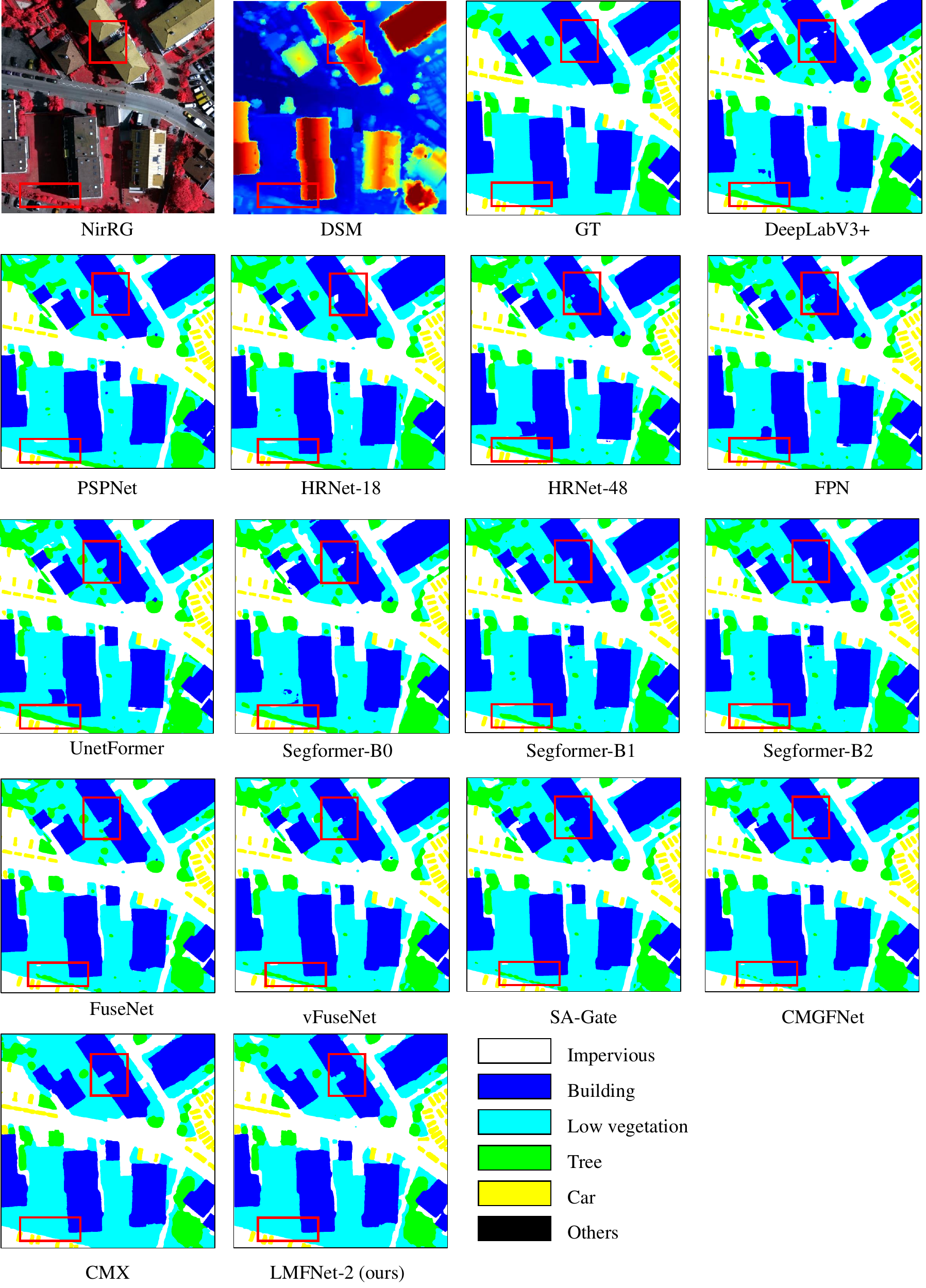}
    \caption{Qualitative comparison on the Vaihingen dataset between ours and other SOTA methods.  The red boxes in the figure mark the areas of focus.}
    \label{fig:vaidemo}
\end{figure}

In the comparison of qualitative results, as shown in Figure \ref{fig:vaidemo}, without DSM data, low vegetation, especially shrubs (the red rectangle at the bottom of the results), are more likely to be mistakenly classified as trees. After adding DSM data, correct classification is achieved. In the case of only having two-dimensional images, building edges might be difficult to accurately identify due to resolution limitations or backgrounds that are close in color to the buildings (the red square at the top of the results). By incorporating the information from DSM, it is possible to clearly distinguish buildings from their surroundings, thereby obtaining visually better building contours.

\subsubsection{Analysis of Model Parameters}




The statistics of the model parameters are shown in the \textit{params} column of Table~\ref{tab:us3d}. It can be observed that in terms of the number of parameters, our method significantly outperforms other methods. This is because we adopt a weight sharing multi-branch network as the backbone. The CMX also adopts MiT as the backbone choice, but in network design, it chose to use two different backbones for feature extraction, resulting in a higher number of parameters than our method. Other multimodal fusion methods also share similar issues with CMX. In contrast, our method has the advantage of being easily scalable; it does not require modifications to the network structure after increasing the number of modalities, and compared to LMFNet-2 ($3.72M$), the parameter count of LMFNet-3 ($4.22M$) only increases by $0.5M$.


    

\section{Discussion}
\label{sec:discusiion}

Compared with existing models, our method demonstrates excellent performance on US3D, ISPRS Potsdam and ISPRS Vaihingen datasets, especially when extra modal data is introduced. This validates the effectiveness of the LMFNet in enhancing the model's capability to process multimodal data. Simultaneously, adopting a weight-shared multi-branch backbone network not only keeps the network lightweight but also maintains high accuracy, proving that efficient network structure is equally important when designing multimodal data fusion models.


%
\subsection{Analysis of the Three Multimodal Feature Merge Operations}

\begin{figure}[ht]
    \centering
    \includegraphics[width=0.78\linewidth]{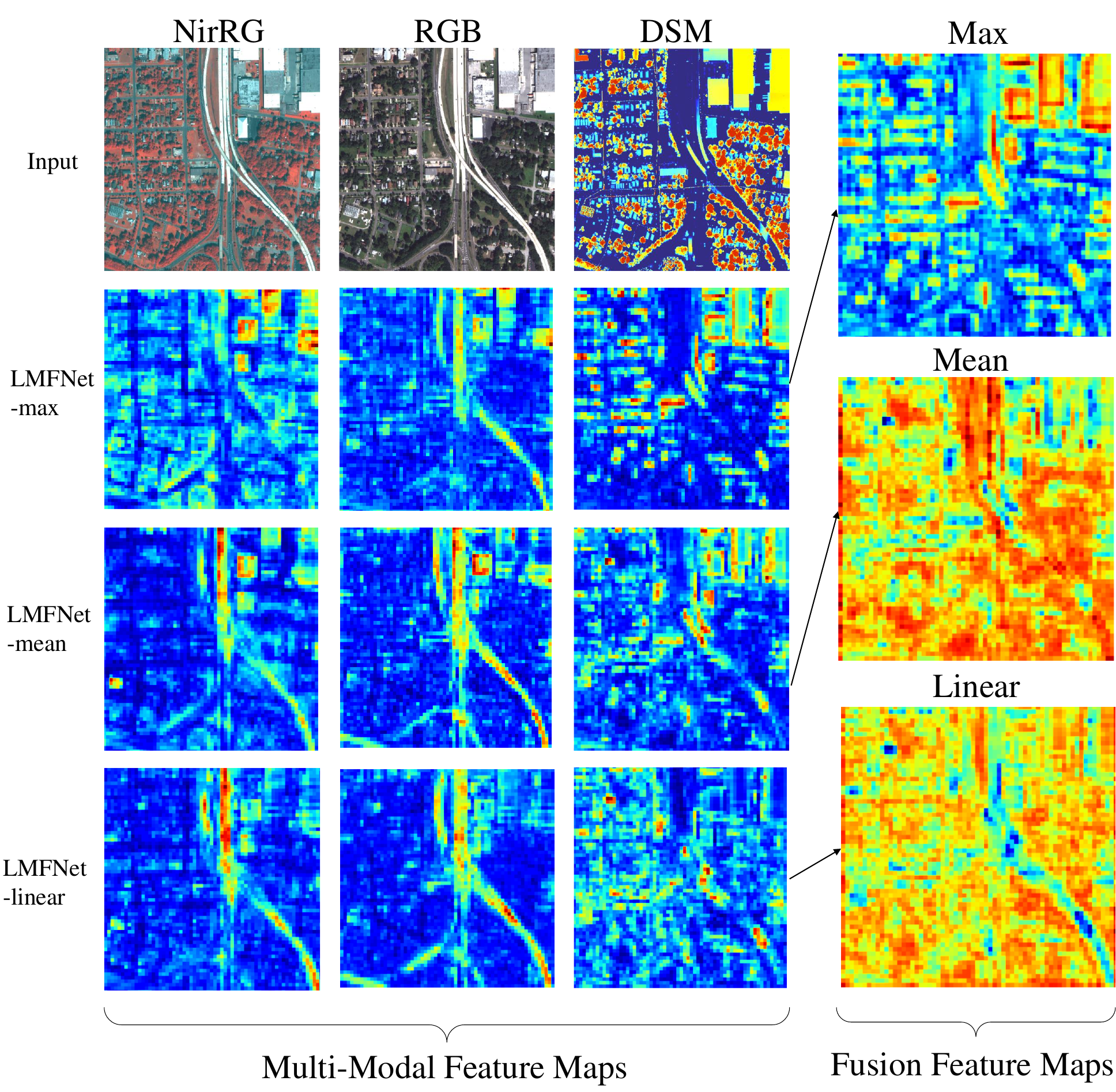}
    \caption{Visualization of multimodal features. The first row in the figure represents the input images, and the rows $a,b,c$ respectively represent the multimodal feature maps using $max$, $mean$, and $linear$ operations. The first three columns show the data before fusion, and the fourth column shows the feature map after fusion.}
    \label{fig:demo1}
\end{figure}

In Table~\ref{tab:meanmax}, we conduct a quantitative assessment of the three fusion methods: $max$, $mean$, and $linear$. To further analyze why the $max$ operation has the best effect, we visualize the features of multimodals before the feature fusion module and the fusion feature maps $F_{out}$ (in Figure~\ref{fig:fusion_module}), as shown in the Figure~\ref{fig:demo1}. From the figure, it can be seen that the input of each modality has a strong response to different areas respectively. The response of NirRG to vegetation is stronger than that to other categories, while RGB responds more strongly to areas with high reflectivity such as roads. DSM shows more significant responses to buildings and bridges with clear elevation information. The features after fusion respond more strongly to the whole. This indicates that our fusion method indeed integrates the advantages of each modality.

Each merge operation has a different logic in the way it integrates information between modalities: 

\textbf{Max}: This operation significantly improves the feature discriminability of multimodal data by selecting the maximum value among all modal features as the fused feature value. The feature map merged by the max operation has better distinguishability on the boundaries, which indicates that the max method can effectively extract the features that contribute the most to the target classification from each modality, thereby enhancing the model's ability to recognize different objects. 

\textbf{Mean}: It calculates the average value of all modal features as the feature for fusion. Compared to the max method, the feature map fused by the mean method shows a strong similarity between the NirRG and RGB modalities, making it difficult for the model to distinguish between these two modalities. This may be because averaging tends to reduce the differences between features, leading to a decrease in model performance in distinguishing features from different modalities. 

\textbf{Linear}: It is a merge operation of linear combination, which also fails to show sufficient distinguishability between NirRG and RGB modalities. This indicates that a simple linear combination may not effectively capture the complex relations between modalities, having limited effectiveness in improving distinguishability between modalities.

In summary, through the comparison of different merge operation, the $max$ performs better in the merging of multimodal data. It can effectively select the most beneficial features for the current task from multiple modalities, thereby improving the model's performance. Additionally, each data modality, NirRG for vegetation, RGB for roads, DSM for buildings and bridges, responds differently to different objects, indicating that before fusion, data from different modalities already contain information that is more sensitive to specific categories. By effective fusion, these advantages can be integrated, making the fused features more generalized and distinguishable, thereby achieving better results in multimodal learning tasks.


\subsection{Analysis of the Roles of the MFFR Layer and the MFSAF Layer in the LMFNet}

\begin{figure}[ht]
    \centering
    \includegraphics[width=0.75\linewidth]{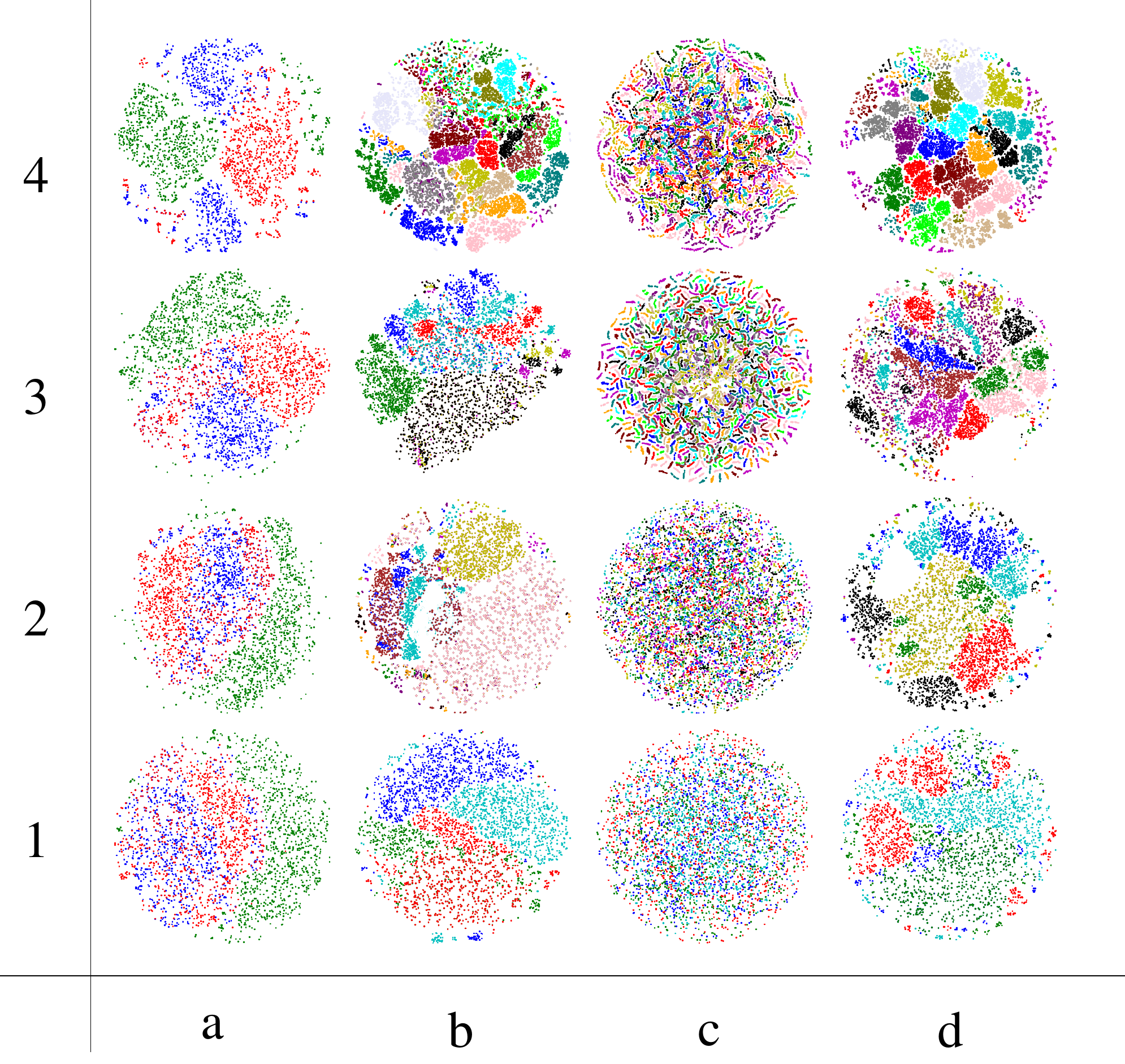}
    \caption{Multimodal feature tsne visualization. The number $n$ on the left represents the output from the $n$-th transformer block. The $a$, $b$, $c$, $d$ below respectively represent the $F(B,C,h,w,N)$, the output of the first MFFR Layer, the output of the MFSAF Layer, and the output of the Add layer in Figure~\ref{fig:fusion_module}.}
    \label{fig:tsne}
\end{figure}

In order to explore the roles of the MFFR Layer and MFSAF Layer, we extract features from multiple scales and fusion stages for visualization to intuitively understand the fusion process. The visualization method used is t-distributed Stochastic Neighbor Embedding (t-SNE)~\cite{van2008visualizing}, which reduces the dimensions of the features to two-dimensional. We use different colors to represent different modalities, as shown in Figure \ref{fig:tsne}.


From the first row and $a$ col of Figure~\ref{fig:tsne}, it can be seen that at the initial stage of multimodal data processing (such as RGB images and NirRG data), the features between different modalities might be intermixed, making them difficult to distinguish. This indicates that despite the different data sources, their characteristics appear quite similar at the initial stages. As the feature extraction progresses to deeper levels (the rows $1$ to $4$  in \ref{fig:tsne}), the characteristics between different modalities start to exhibit more distinguishable properties. By the time of the fourth layer of features, these distinguishable characteristics become evident.

\textbf{The role of the MFFR Layer}: by introducing the first MFFR Layer, the distinction among features from different modalities has been significantly enhanced, as shown in the $a$ and $b$ cols of Figure~\ref{fig:tsne}. The MFFR layer mainly focuses on reconstructing the feature space, uncovering more independent and distinctly distinguishable features.

\textbf{The role of the MFSAF Layer}: as shown in the $c$ col of the Figure~\ref{fig:tsne}, the introduction of the MFSAF Layer enables the model to further enhance the fusion of features between modalities within an already well-differentiated feature space. At this point, although the MFFR layer has already promoted the independence between features, the MFSAF Layer introduces a level of mixing on this basis to strengthen the depth and effectiveness of feature fusion.

As the $d$ col of the Figure~\ref{fig:tsne}, after processing by the MFSAF Layer, the model undergoes further reconstruction of features through another MFFR Layer. This helps to more clearly delineate the boundaries between different modalities on the basis of previous fusion, especially in terms of shallow features. The entire process achieves effective multimodal feature fusion, and thereby significantly enhances the accuracy of classification, through reconstructing, mixing, and then reconstructing the features again.
In summary, the research findings of this paper provide valuable insights into the deep learning processing and fusion of multimodal remote sensing data, laying a solid foundation for future research in this field.

\section{Conclusion}
\label{sec:con}

In this work, we have developed the \textbf{L}ightweight \textbf{M}ultimodal \textbf{F}usion \textbf{Net}work, \textbf{LMFNet}, an innovative approach that efficiently integrates high-resolution multispectral imagery and Digital Surface Models (DSM) for semantic segmentation tasks in remote sensing. In LMFNet, we introduce a novel module for multimodal fusion, which reconstructs and fuses  multimodal data in a hidden space of multimodal information, using a tensor-based perspective. This enables the seamless integration of an arbitrarily number of modal types. Specifically, our proposed MFFR Layer maps multimodal features into the hidden space for feature reconstruction, while the MFSAF Layer realizes the fusion of multimodal features within this hidden space. Additionally, to minimize the gap between DSM and multi-spectral data, we also employ a colormap method to map single-band DSM into rgb space. Extensive experiments conducted on the US3D dataset have identified the optimal parameter combination for our network. The results on the US3D, ISPRS Potsdam, and ISPRS Vaihingen datasets validate the effectiveness of LMFNet, which achieves the optimal performance when fully utilizing data from all three modalities, compared to other state-of-the-art methods.

In the future, much work can be carried out regarding the exploration of multimodal information. In our work, we preliminarily demonstrate that employing a powerful backbone can simultaneously extract features from multiple modalities. However, the data fusion part still requires a large number of annotated samples for training. How to fully utilize the current multimodal remote sensing data remains one of the research questions worth investigating. As visual large models like Segment Anything (SAM)~\cite{kirillov_segment_2023} continue to emerge, how to apply them in multimodal data to achieve multimodal fusion segmentation without the need for retraining will also be the focus of our future research work.

\section{Funding}
This research was supported in part by the National Natural Science Foundation of China (No. 42101346), in part by Science and Technology Commission of Shanghai Municipality (No. 22DZ1100800), in part by the ``Unveiling and Commandin'' project in the Wuhan East Lake High-tech Development Zone (No. 2023KJB212), and in part by the China Postdoctoral Science Foundation (No. 2020M680109).

\section{Author Contribution}
Tong Wang: Conceptualization, Methodology, Software, Writing - Original Draft. Guan-zhou Chen: Resources, Data Curation, Writing - Review \& Editing. Xiaodong Zhang: Supervision, Project administration, Funding acquisition. Chenxi Liu: Writing - Review \& Editing, Investigation. Jiaqi Wang: Validation. Xiaoliang Tan: Validation. Wenlin Zhou: Visualization. Chanjuan He: Visualization.

\appendix
\section{Declaration of Competing Interest}
The authors declare that they have no known competing financial interests or personal relationships that could have appeared to influence the work reported in this paper.

\section{Comparions of Different Rendering Methods for Single-channel-modal Data}
\label{sec:colormaps}

\begin{figure}[htb]
     \centering
     \includegraphics[width=0.5\linewidth]{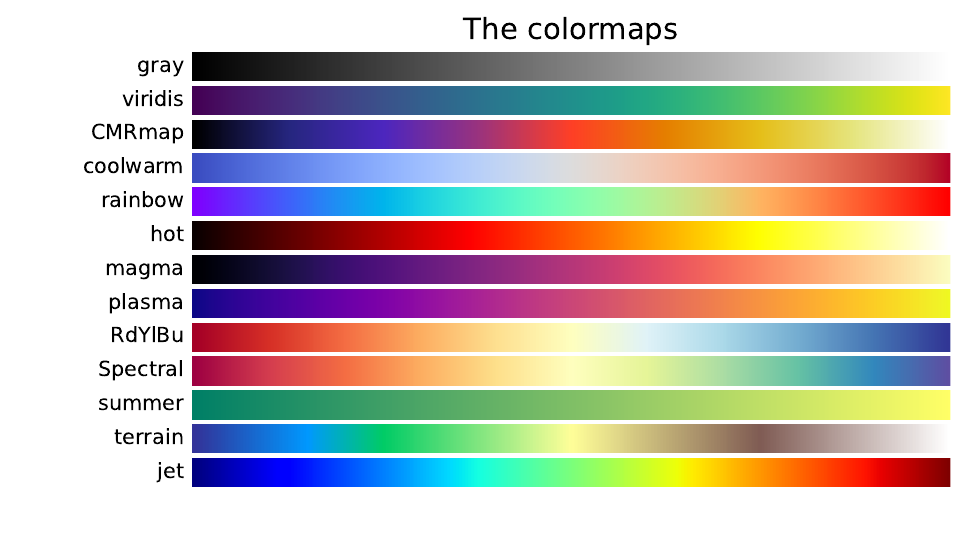}
     \caption{ Colormaps in Rendering Methods for Single-channel-modal Data. }
     \label{fig:colormap}
 \end{figure}

There are currently various colormap methods available.  In our work, we select several commonly used color mapping methods, as shown in Figure~\ref{fig:colormap}. We test different colormap methods to find the optimal representation for DSM data. The specific methods used and the quantitative results are shown in the Table~\ref{tab:colormap}.

\begin{table}[ht]
 \caption{Different Colormap of DSM results on US3D dataset. The blod is the best.}
\begin{adjustbox}{width=1\columnwidth,center}

    \centering
    
    \begin{tabular}{ccccccccc} 
     \hline
     \multirow{2}{*}{Methods} & \multicolumn{5}{c}{$IoU$ (\%)}   &  \multirow{2}{*}{$mF1$ (\%)}&  \multirow{2}{*}{$mIoU$ (\%)}&  \multirow{2}{*}{$OA$ (\%)}\\
     \cline{2-6}
          &  Ground&  Foliage&  Building&  Water&  Bridge&   & &\\ 
         \hline
         gray &  92.24 &	78.69 &	80.02 &	84.07 &	85.06 &	91.24 &	84.02&	93.23 \\
viridis  & 92.31          & 78.97          & 81.13          & 85.78          & 87.06          & 92.01          & 85.05          & 93.53          \\
CMRmap   & 92.03          & 78.63          & 80.64          & \textbf{85.94} & 86.87          & 91.86          & 84.82          & 93.42          \\
coolwarm & \textbf{92.63} & 78.94          & \textbf{81.08} & 85.39          & 87.28          & 91.82          & 85.06          & \textbf{93.54} \\
rainbow  & 91.66          & 79.14          & 80.89          & 85.23          & 87.41          & 91.68          & 84.87          & 93.44          \\
hot      & 91.61          & 79.33          & 80.81          & 85.66          & 86.52          & 91.83          & 84.79          & 93.41          \\
magma    & 91.41          & 79.08          & 80.21          & 84.68          & 87.16          & 91.64          & 84.51          & 93.28          \\
plasma   & 91.94          & \textbf{79.98} & 79.91          & 84.87          & 87.08          & 91.81          & 84.76          & 93.39          \\
RdYlBu   & 91.88          & 78.98          & 80.61          & 85.28          & 87.63          & 91.89          & 84.88          & 93.45          \\
Spectral & 92.08          & 79.04          & 80.29          & 85.08          & 87.71          & 91.87          & 84.84          & 93.43          \\
summer   & 91.9           & 79.21          & 80.17          & 84.89          & 87.23          & 91.76          & 84.68          & 93.36          \\
terrain  & 92.26          & 79.82          & 80.79          & 84.89          & \textbf{87.54} & 91.72          & 85.06          & 93.53          \\
jet      & 92.61          & 79.52          & 80.83          & 85.61          & 86.90          & \textbf{91.88} & \textbf{85.09} & 93.53         
\\\hline
    \end{tabular}

   \end{adjustbox}
    \label{tab:colormap}
\end{table}
%
It can be observed that compared to the gray method, the accuracy after adopting color rendering has experienced a slight improvement. Different colormap methods have a significant impact on the model\'s performance in specific categories, indicating that the strategy of color rendering in DSM data processing needs to be meticulously selected according to the target task. Specifically, we found that the coolwarm colormap shows the best accuracy in processing Ground and Building categories, while the plasma colormap performs better in the Foliage category. This difference may stem from different colormap schemes emphasizing image characteristics in various visual ways, thereby affecting the model's ability to learn and recognize these features. Although different colormaps have their advantages in specific tasks, overall, the jet colormap has been selected as the most suitable scheme due to its relatively balanced performance across all categories. Its balanced performance across multiple classification tasks means it contributes most significantly to the overall improvement of the model's performance, especially when considering the $mIoU$ and average $mF1$.

\section{Acknowledgement}

The numerical calculations in this paper have been done on the supercomputing system in the Supercomputing Center of Wuhan University. Tribute to open-source contributors, and gratitude to the authors of open-source works such as PyTorch, GDAL, Open-MMLab, MMSegmentation, and others.

 \bibliographystyle{elsarticle-num-names} 
 \bibliography{main}





\end{document}